\documentclass[10pt,journal,compsoc]{IEEEtran}
\usepackage{ifpdf}

\ifCLASSOPTIONcompsoc
  \usepackage[nocompress]{cite}
\else
  \usepackage{cite}
\fi
\ifCLASSINFOpdf
   \usepackage[pdftex]{graphicx}
   \graphicspath{{./figures/}{./images/}}
\else
\fi
\usepackage{url}

\usepackage{color}
\usepackage{cite}
\usepackage{amsmath,amssymb,amsfonts}
\usepackage{algorithmic}
\usepackage{graphicx}
\usepackage{textcomp}
\usepackage{xcolor}
\usepackage{booktabs}
\usepackage{multirow}
\usepackage{times}
\usepackage{epsfig}
\usepackage{graphicx}
\usepackage{amsmath}
\usepackage{mathrsfs}
\usepackage{amssymb}
\usepackage{float}
\usepackage[section]{placeins}
\usepackage[pagebackref=true,breaklinks=true,letterpaper=true,colorlinks,bookmarks=false]{hyperref}
\usepackage{bbding}

\usepackage{tabu}

\begin{document}
%
\title{Adaptive Multi-view and Temporal Fusing Transformer for 3D Human Pose Estimation}
%
%
%
%

\author{Hui Shuai*, Lele Wu*,
        and~Qingshan Liu,~\IEEEmembership{Senior Member,~IEEE}
\IEEEcompsocitemizethanks{
\IEEEcompsocthanksitem The authors are with Engineering Research Center of Digital Forensics, Ministry of Education, School of Computer and Software, Nanjing University of Information Science and Technology, Nanjing, 210044, China. (E-mail: huishuai13@nuist.edu.cn, llwu@nuist.edu.cn, qsliu@nuist.edu.cn)
\IEEEcompsocthanksitem Hui Shuai and Lele Wu equally contributed on the work. Qingshan Liu is the corresponding author.
Code is available in \url{https://github.com/lelexx/MTF-Transformer}.}
\thanks{Manuscript received 03 Nov. 2021; revised 18 Apr. 2022 and 19 Jun. 2022; accepted 29 Jun. 2022}}

%
%

\markboth{Journal of \LaTeX\ Class Files,~Vol.~14, No.~8, August~2015}%
{Shell \MakeLowercase{\textit{et al.}}: Bare Demo of IEEEtran.cls for Computer Society Journals}
%



\IEEEtitleabstractindextext{%
\begin{abstract}
This paper proposes a unified framework dubbed Multi-view and Temporal Fusing Transformer (MTF-Transformer) to adaptively handle varying view numbers and video length without camera calibration in 3D Human Pose Estimation (HPE).
It consists of Feature Extractor, Multi-view Fusing Transformer (MFT), and Temporal Fusing Transformer (TFT).
Feature Extractor estimates 2D pose from each image and fuses the prediction according to the confidence.
It provides pose-focused feature embedding and makes subsequent modules computationally lightweight.
MFT fuses the features of a varying number of views with a novel Relative-Attention block.
It adaptively measures the implicit relative relationship between each pair of views and reconstructs more informative features.
TFT aggregates the features of the whole sequence and predicts 3D pose via a transformer.
It adaptively deals with the video of arbitrary length and fully unitizes the temporal information.
The migration of transformers enables our model to learn spatial geometry better and preserve robustness for varying application scenarios.
We report quantitative and qualitative results on the Human3.6M, TotalCapture, and KTH Multiview Football II.
Compared with state-of-the-art methods with camera parameters, MTF-Transformer obtains competitive results and generalizes well to dynamic capture with an arbitrary number of unseen views.
\end{abstract}

}

\maketitle

\IEEEdisplaynontitleabstractindextext

%
\IEEEpeerreviewmaketitle

\IEEEraisesectionheading{\section{Introduction}\label{sec:introduction}}

%
%
%
%


\IEEEPARstart{T}{hree}-dimensional human pose estimation (HPE) aims to predict 3D human pose information from images or videos,
in which skeleton joint location is the primary output result to carry pose information.
It plays a fundamental role in many applications, such as action recognition~\cite{Wang_2016_CVPR, liu2017skeleton,luvizon2020multi},
human body reconstruction~\cite{Guler_2019_CVPR,fang2018learning}, and robotics manipulation~\cite{ehlers2016human,tao2020trajectory}.

With the emergence of deep learning, 3D HPE has made considerable progress. Especially, 2D-to-3D~\cite{martinez2017simple,zhang2020learning,wang2018robust,9174911} methods
have superior performance owing to intermediate 2D supervision\cite{liu2021recent}.
In practice, the 2D-to-3D pipeline involves several variable factors deriving from different application scenarios,
including the number of views, the length of the video sequence, and whether using camera calibration.

In the monocular scene, most works~\cite{martinez2017simple, moreno20173d, sharma2019monocular, zhao2019semantic}
estimate body structure from a static image with elaborate networks such as Convolutional Neural Networks and Graph Convolutional Networks.
This scheme is convenient since a single image is easy to obtain and process.
Nevertheless, the information in a single image is insufficient considering the occlusion and depth ambiguity.
For compensation, some works~\cite{pavllo20193d, cai2019exploiting, wang2020motion, zeng2021learning, zheng20213d} utilize temporal information from video sequences.
Sequential variation in the video is conducive to revealing the human body's structure.
However, continuous images contain more homogeneous information rather than complementary clues.
In a word, monocular 3D HPE is convenient to implement, but recovering 3D structure from 2D images is always an ill-posed problem.

Recently, prevalent works~\cite{he2020epipolar, iskakov2019learnable, zhang2021adafuse, qiu2019cross, ma2021transfusion} tend to utilize multi-view geometric constraints.
Most existing multi-view methods aggregate information from different views via projective geometry, depending on calibrated camera parameters.
Camera parameters incorporate solid prior knowledge into the network but are difficult to calibrate accurately in dynamic capture.
To this end, some works \cite{huang2020deepfuse} attempt to fuse multi-view features without calibration,
but they have strict requirements on camera configuration and the number of views.
In addition, massive computation in the geometric space hinders multi-view methods to deal with video sequences.
Overall, most existing multi-view methods are more accurate than monocular methods,
but camera calibration and computation overhead limit their application scenarios.

Each method, as mentioned above, targets one or a few particular combinations of those variable factors and is not compatible with others,
limiting the flexibility of the 3D HPE algorithm.
Thus, developing a unified framework that can adaptively handle all the factors is essential.
The main obstacles are that
(1) Most deep learning modules, such as fully connected layers, long and short-term memory (LSTM), and GCN,
are not friendly to variable-length input.
Moreover, these modules still have generalization problems even with careful adjustments to handle variable-length input.
(2) Most methods rely on camera calibration to deal with multi-view information,
but precise camera parameters are unrealistic to calibrate synchronously in dynamic capture.
(3) Some methods are too computationally expensive to deal with multi-view videos.
Accordingly, a unified framework needs to be compatible with monocular to multi-view, single-image to videos 3D HPE:
(1) It should effectively integrate an arbitrary number of multi-view features in uncalibrated scenarios.
(2) It should adaptively fuse temporal features in the variable-length videos and be compatible with a single image.
(3) It should be lightweight enough and have generalization capability.

To satisfy these requirements, we propose a unified framework to deal with variable multi-view sequences without calibration,
named Multi-view and Temporal Fusing Transformer (MTF-Transformer) because the transformer can perceive the global relationship of a varying number of tokens and aggregate them adaptively \cite{vaswani2017attention}.
MTF-Transformer consists of Feature Extractor, Multi-view Fusion Transformer (MFT), and Temporal Fusion Transformer (TFT).
In the Feature Extractor, a pre-trained 2D detector predicts the 2D pose of each frame first. Then, coordinates and confidence are encoded into a vector via a feature embedding module,
discarding the image features. It makes subsequent modules lightweight and focuses on lifting the 2D pose into the 3D pose.
MFT is designed to fuse the features of multiple views into more informative ones.
It integrates the relationship between the views into the procedure that calculates the key, query, and value in the Relative-Attention block, avoiding camera calibration.
In TFT, we employ a conventional transformer to capture temporal information.
It is worth mentioning that, to make the MTF-Transformer adaptive to the input of an arbitrary number of views and length of sequences,
we design a random mask mechanism in both MFT and TFT, referring to the dropout mechanism \cite{tinto1975dropout}.

We evaluate our method on Human3.6M~\cite{ionescu2013human3}, TotalCapture~\cite{joo2018total}, and KTH Multiview Football II~\cite{kazemi2013multi} quantitatively and qualitatively.
We also conduct detailed ablation study experiments to verify the effectiveness of each module.
Experiment results demonstrate that MTF-Transformer outperforms camera parameter-free methods.
Besides, MTF-Transformer can be directly applied to the scenarios with different configurations from the training stage, bridging the generalization gap significantly.
In short, our contributions are:
\begin{itemize}
  \item We proposed a unified framework (MTF-Transformer) for 3D HPE. It is adaptive to scenarios with videos of arbitrary length and from arbitrary views without retraining.
  \item We design a novel Multi-view Fusing Transformer (MFT), where the relationship between views is integrated into the Relative-Attention block.
  MFT reconstructs the features from multiple views according to the estimated implicit relationship, avoiding the need for camera calibration.
  \item We introduce the random mask mechanism into MFT and Temporal Fusing Transformer (TFT) to make them robust to variable view number and video length.
  \item Not only does our model outperform camera parameter-free models in precision, but it also has a better generalization capability to handle diverse application scenarios.
\end{itemize}

%

\section{Related Work}
This section firstly summarizes 3D human pose estimation works, including monocular and multi-view methods.
Then, we review the transformer technology and introduce the methods that apply the transformer in 3D human pose estimation and some other related tasks.

\subsection{3D Human Pose Estimation}
Fundamentally, 3D HPE is to reconstruct the 3D body structure from 2D data.
It is an ill-posed inverse task as one 2D image corresponds to many possible 3D poses, further amplified by occlusions, background clutters, etc.
Thus, utilizing all kinds of clues, such as the mutual constraint between the joints in the image, the complementary information in videos, and the spatial geometric relationship from multiple viewpoints, to piece together the most likely 3D pose is the rationale for 3D HPE.
According to the different clues, 3D HPE methods are divided into categories and developed into several frameworks that handle specific application scenarios.

\subsubsection{Monocular 3D Human Pose Estimation}
With the pattern self-organizing and non-linear mapping capacity of deep neural networks,
many approaches~\cite{sun2018integral,pavlakos2017coarse,zhou2019hemlets,li20143d,kanazawa2018end,wang2020predicting, martinez2017simple, li2020cascaded} directly map pixel intensities to 3D poses from a single image.
It forces DNNs to remember the pattern and infer the 3D pose.
These networks are difficult to learn and rely on tremendous labeled samples, resulting in unsatisfactory performance and generalization capability.
Therefore, prior constraints between joints are utilized to determine the special pose.
Fang~\cite{fang2018learning} et al. incorporate kinematics, symmetry, and motor coordination grammar in 3D pose estimation.
Some works employ GCN to model the constraints between the joints~\cite{liu2020semi,zhao2019semantic}.
These methods are devoted to digging into the image's potential information, but such a manner is insufficient to solve an ill-posed problem.
To solve the ambiguity of a single image, more works~\cite{cai2019exploiting,shi2020motionet, wu2021limb, xu2020deep, 2019Trajectory} pay attention to temporal consistency in the video.
For example, Pavllo et al. \cite{pavllo20193d} transform a sequence of 2D poses through temporal convolutions.
Cai et al. propose a graph-based method to incorporate spatial dependencies and temporal consistences\cite{cai2019exploiting}.
Wang et al.~\cite{shi2020motionet} employ a novel objective function to involve motion modeling in learning explicitly.
Temporal information compensates for the incompleteness of 3D geometry, improving the performance of 3D HPE.
In general, monocular methods are easy to implement as there is no need for camera calibration.
However, to piece up the 3D structure from 2D images,  it is evident that the clues from multiple viewpoints are better alternatives.

\begin{figure*}[ht!]
	\centerline{\includegraphics[width=\textwidth]{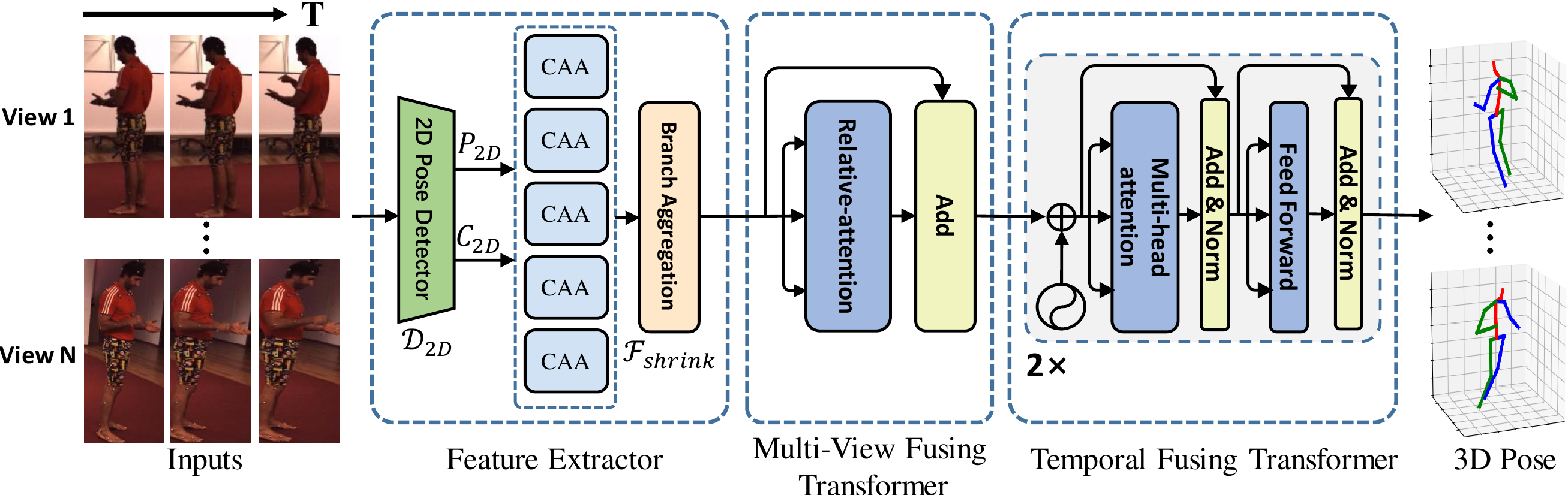}}
	\caption{The architecture of MTF-Transformer. It consists of three successive modules: Feature Extractor, Multi-view Fusing Transformer (MFT), and Temporal Fusing Transformer (TFT). Feature Extractor predicts 2D pose ($P_{2D}$ and $C_{2D}$) first and then encodes 2D pose into a feature vector for each frame. MFT measures the implicit relationship between each pair of views to reconstruct the feature adaptively. TFT aggregates the temporal information of the whole sequence and predicts the 3D pose of the center frame.}
	\label{fig_pipeline}
\end{figure*}

\subsubsection{Multi-view 3D Human Pose Estimation}
To tackle the occlusion and depth ambiguity, multi-view methods~\cite{dong2019fast, qiu2019cross, kadkhodamohammadi2021generalizable, he2020epipolar, iskakov2019learnable} exploit geometric information from multiple views to infer 3D pose.
Most utilize intrinsic and extrinsic camera parameters to fuse 2D features from different views.
For example,
He et al.~\cite{he2020epipolar} aggregate features on epipolar lines between different views, depending on camera parameters in specific camera configurations.
Iskakov et al. \cite{iskakov2019learnable} utilize volumetric grids to fuse features from different views with camera parameters and regress root-centered 3D pose through a learnable 3D CNN.
Despite predicting 3D poses reliably, volumetric approaches are computationally demanding.
These methods require precise camera parameters but can not generalize to scenarios with new camera configurations, not to mention the dynamic capture.
Huang et al. \cite{huang2020deepfuse} propose a new vision-IMU data fusion technique to avoid strict requirements on camera configuration and the number of views.
FLEX \cite{gordon2021flex} introduce to predict joint angles and bone lengths invariant to the camera position rather than directly 3D positions, so calibration is obviated.
Nevertheless, it is complicated, and its performance degenerates with only a few views.
Multi-view pose estimation methods are more accurate due to adequate feature fusing via projective geometry. However, another side of the coin is that these methods rely on the restricted camera configuration explicitly or implicitly, limiting their application scene.
\\
\\
Monocular and multi-view methods exploit the clues from different aspects and fit particular application scenarios.
Unlike these methods, we attempt to fuse all the clues adaptively in a unified network that can predict robust 3D poses in all the application scenarios.
So, the critical component is to find a mechanism that organically integrates the information from different aspects.

\subsection{Transformer in 3D Pose Estimation}
Transformer and self-attention have tremendously succeeded in Natural Language Processing and Computer Vision~\cite{raghu2021vision}.
The self-attention module can adaptively capture long-range dependencies and global correlations from the data.
In 3D pose estimation, the core is to integrate the information from spatial 2D joints, temporal sequence, and multiple viewpoints.
Thus, the transformer is suitable for handling these aspects of information, and some works utilizing the transformer models have recently emerged.

Following the line of lifting 2D to 3D, some monocular methods improve the performance by introducing the transformer.
Among them, METRO~\cite{lin2021end} employs a multi-layer transformer architecture with progressive dimensionality reduction to regress the 3D coordinates of the joints and vertices.
PoseGTAC~\cite{ijcai2021-188} proposes graph atrous convolution and graph transformer layer to extract local multi-scale and global long-range information, respectively.
 More works handle spatial-temporal clues with the transformer to alleviate occlusion and depth ambiguity in a single image.
For example, LiftFormer~\cite{llopart2020liftformer} estimates 3D pose from a sequence of 2D keypoints with self-attention on long-term information. MHFormer~\cite{li2021mhformer} proposes a Multi-Hypothesis Transformer (MHFormer) to learn spatio-temporal representations of multiple plausible pose hypotheses and aggregates the multi-hypothesis into the final 3D pose.
Strided Transformer~\cite{li2022exploiting} incorporates the strided convolution into the transformer to aggregate long-range information in a hierarchical architecture at low computation.
PoseFormer~\cite{zheng20213d} proposes a purely transformer-based approach to model the spatial relationships between 2D joints and temporal information in videos.
Naturally, the transformer is also used to aggregate the multi-view clues, but it usually works with epipolar geometric while camera parameters are essential prerequisites.
Epipolar transformer~\cite{he2020epipolar} leverages the transformer to find the point-point correspondence in the epipolar line.
TransFusion~\cite{ma2021transfusion} further proposes the concept of epipolar field to encode 3D positional information into the transformer.

This tendency demonstrates the potential of the transformer for feature fusing in 3D pose estimation.
Moreover, the transformer is inherently adaptive to a variable number of input tokens.
Thus, our concerns focus on generalization capability and calibration avoidance.
Fortunately, the transformer generalizes well to the configurations different from the training phase in some other tasks.
For example, Pooling-based Vision Transformer~\cite{heo2021rethinking} improves model capability and generalization performance via designing a pooling layer in ViT.
Neural Human Performer~\cite{kwon2021neural} synthesizes a free-viewpoint video of an arbitrary human performance, and it generalizes to unseen motions
and characters at test time.
It adaptively aggregates multi-time and multi-view information with temporal and multi-view transformer.
However, Neural Human Performer fuses the multi-view features that are pixel-wisely matched by a parametric 3D body model (SMPL).
Such pixel-pixel correspondence between multiple viewpoints is stronger than pixel-epipolar correspondence.
So, fusing multi-view features without camera calibration remains an open problem for us.

\section{Method}
The purpose of our framework is to adaptively handle features from an arbitrary number of views and arbitrary sequence length without camera calibration.
As shown in \figurename~\ref{fig_pipeline}, the basic idea is to embed 2D detections into vectors first, then fuse multi-view features, and finally aggregate temporal clues to predict 3D joints. This framework consists of Feature Extractor, Multi-view Fusing Transformer (MFT), and Temporal Fusing Transformer (TFT).

\subsection{Feature Extractor}
Feature Extractor uses a pre-trained 2D pose detector (e.g., CPN~\cite{chen2018cascaded}) to obtain 2D predictions and then maps them into 1D feature vectors through a feature embedding module.

\begin{figure}[t]
	\centerline{\includegraphics[width=0.45\textwidth]{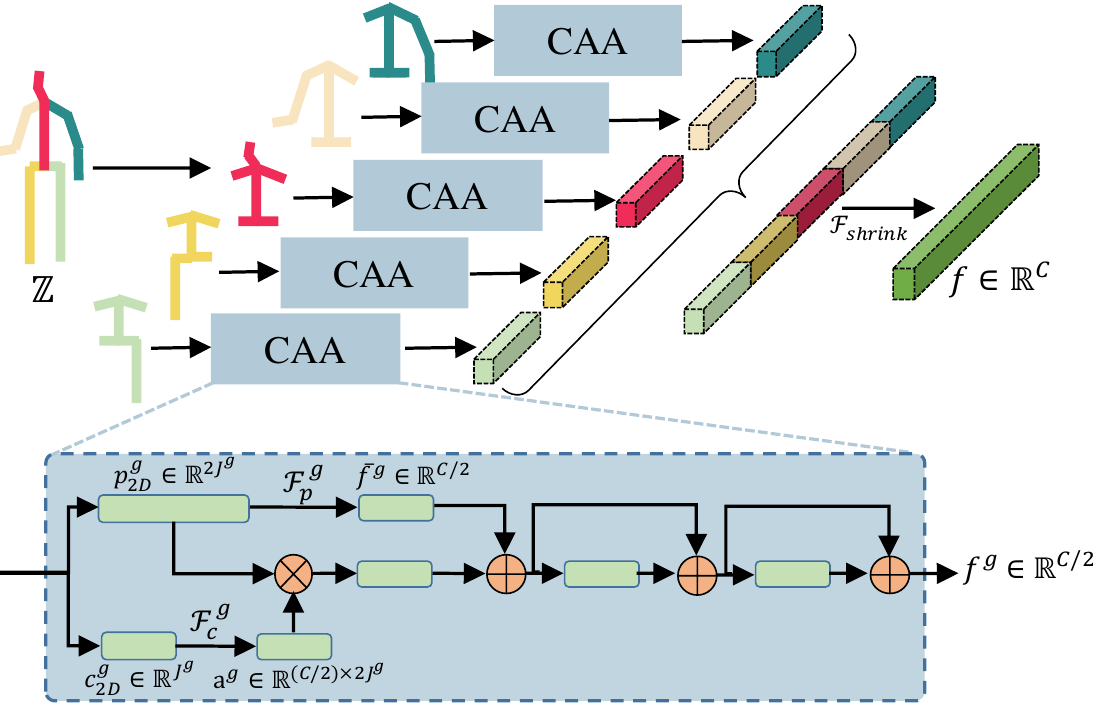}}
	\caption{The feature embedding module encodes the 2D prediction into a feature vector. It splits the 2D prediction into five partitions and then uses five branches to extract features. Finally, the features of five partitions are concatenated and mapped to a global feature $\boldsymbol{f}$.}
	\label{fig_feature_extract}
\end{figure}

Taking multi-view sequences $\mathbf{\mathcal{I}}=\{\mathbf{I_i}\}_{i=1}^{N\times T}$ with $N$ views and $T$ frames as input, each frame is an image $\mathbf{I}\in \mathbb{R}^{W\times H\times 3}$. As the following operations are conducted on each frame, we omit $N$ and $T$ for simplicity here.
For each frame, Feature Extractor first uses a pre-trained 2D pose detector $\mathcal{D}_\text{2D}$ to infer the 2D prediction:
\begin{equation}
  \mathbb{Z} = \mathcal{D}_\text{2D}\left(\mathbf{I}\right)
\end{equation}
where $\mathbb{Z}=\{\mathbf{P}_\text{2D}, \mathbf{C}_\text{2D}\}$, $\mathbf{P}_\text{2D}=\{\boldsymbol{p}_{j}\}_{j=1}^{J}$ represents $J$ coordinates of the 2D pose and
$\mathbf{C}_\text{2D}=\{c_{j}\}_{j=1}^{J}$ represents the confidence of these coordinates.
Then a feature embedding module encodes the predicted 2D pose into a feature vector (as shown in \figurename~\ref{fig_feature_extract}).
The movements of the limbs and head are relatively independent, so we divide the human body joints into five partitions and deal with them in five parallel branches. The five partitions are the head, left and right arms, and left and right legs:
\begin{align}
  & \mathbf{P}_\text{2D}^{g}=\{\boldsymbol{p}_{k} | k\in \mathbf{S}^{g}\} \\
  & \mathbf{C}_\text{2D}^{g}=\{c_{k} | k\in \mathbf{S}^{g}\}
\end{align}
where $g$ refers to the $g$-th partition, $g\in\{1,2,3,4,5\}$, $\mathbf{P}_\text{2D}^{g}$, $\mathbf{C}_\text{2D}^{g}$ are subset of $\mathbf{P}_\text{2D}$, $\mathbf{C}_\text{2D}$,
$\mathbf{S}^{g}\subset\{1,2,...,J\}$ represents the index set belongs to the $g$-th partition.
For matrix multiplication, $\mathbf{P}_\text{2D}^{g}$, $\mathbf{C}_\text{2D}^{g}$ are reshaped into vectors that $\boldsymbol{p}_\text{2D}^{g}\in\mathbb{R}^{2J^{g}}$, $\boldsymbol{c}_\text{2D}^{g}\in \mathbb{R}^{J^{g}}$.

Since the 2D pose inferred from the pre-trained detector is unreliable due to motion blur and occlusion, simply fusing them may lead to unstable performance.
Previous works, such as FLEX~\cite{gordon2021flex}, directly concatenate the 2D pose and confidence values together for aggregation
but they ignore the effects of unreliable inputs on the features as the pose changes.
In order to alleviate this issue, we utilize the confidence to modulate coordinates. Specifically, Confidence Attentive Aggregation (CAA) extracts local feature $\boldsymbol{f}^{g} \in \mathbb{R}^{C / 2}$ for each part, $C$ is the dimension of the output of Feature Extractor. It can be formulated as:
\begin{align}
  \boldsymbol{\bar{f}}^{g}&=\mathcal{F}_{p}^{g}\left(\boldsymbol{p}_\text{2D}^{g}\right) \\
  \mathbf{a}^{g}&=\mathcal{F}_{c}^{g}\left(\boldsymbol{c}_\text{2D}^{g}\right) \\
  \boldsymbol{f}^{g}&=\mathcal{F}_{res}^{g}\left(\boldsymbol{\bar{f}}^{g}+\mathbf{a}^{g}\cdot\boldsymbol{p}_\text{2D}^{g}\right)
\end{align}
where $\mathcal{F}_{p}^{g}$ is a fully connected layer to map 2D coordinates $\boldsymbol{p}_\text{2D}^{g}$ to initial feature vectors $\boldsymbol{\bar{f}}^{g}\in\mathbb{R}^{C/2}$,
$\mathcal{F}_{c}^{g}$ is another fully connected layer to learn a attention matrix $\mathbf{a}^{g}\in\mathbb{R}^{(C/2)\times 2J^{g}}$ from the confidence $\boldsymbol{c}_\text{2D}^{g}$.
The third fully connected layer $\mathcal{F}_{res}^{g}$ aggregates initial feature vectors $\boldsymbol{\bar{f}}^{g}$ with 2D coordinates $\boldsymbol{p}_\text{2D}^{g}$ modulated by attention matrix $\mathbf{a}^g$. It consists of two res-blocks \cite{martinez2017simple}.

We further concatenate features of five partitions together and map them into a global feature $\boldsymbol{f}\in\mathbb{R}^{C}$. This procedure can be described as:
\begin{align}
	\boldsymbol{f} &= \mathcal{F}_{shrink}\left(\text{Concat}\left(\boldsymbol{f}^{1}, \boldsymbol{f}^{2}, \boldsymbol{f}^{3}, \boldsymbol{f}^{4}, \boldsymbol{f}^{5}\right)\right)
\end{align}
where $\mathcal{F}_{shrink}$ is another fully connected layer. It maps features from five branches to the global feature of each frame.
For the input multi-view sequence $\mathbf{\mathcal{I}}$ with $N\times T$ frames, Feature Extractor extracts the feature $\mathbf{X}\in\mathbb{R}^{C \times N \times T}$
for the subsequent pipeline.

\subsection{Multi-view Fusing Transformer}
We target to measure the relationship between the features from an arbitrary number of views and fuse these features adaptively.
Transformer models are characterized by the ability to model dependencies in the input tokens regardless of their distance and enable immediate aggregation of global information~\cite{vaswani2017attention}.
Thus, the transformer is a candidate to complete this task.
Nevertheless, the conventional transformer does not meet our requirements in position encoding,
and Point Transformer~\cite{zhao2021point} has limitations in manipulating the value item.
So, we design a Relative-Attention that measures the relative relationship between multiple viewpoints and employs a more elaborate value transform procedure.

\subsubsection{Revisit Transformer and Self-attention}
The transformer is a family of models consisting of the self-attention block, appending the position encoding, and the mask block.
The position encoding provides a unique coding for each input token.
The mask block truncates some nonexistent connections based on prior knowledge.
Self-attention operator transforms the input feature vectors $\mathcal{X}=\{\boldsymbol{x}_{i}\}_{i=1}^{N}$
into output feature vectors $\mathcal{Y}=\{\boldsymbol{y}_{i}\}_{i=1}^{N}$, one output feature vector $\boldsymbol{y}_{i}$ is a weighted sum of all the input feature vectors.
Typically, self-attention operators can be classified into scalar attention and vector attention~\cite{zhao2021point}.

The scalar dot-product attention can be formulated as follows:
\begin{equation}
\boldsymbol{y}_{i}=\sum_{\boldsymbol{x}_{j} \in \mathcal{X}} \rho\left(\varphi\left(\boldsymbol{x}_{i}\right)^{\top} \psi\left(\boldsymbol{x}_{j}\right)+\delta\right) \alpha\left(\boldsymbol{x}_{j}\right)
\end{equation}
where $\varphi$, $\psi$, and $\alpha$ are pointwise feature transformations, such as linear projections or MLPs,
$\varphi\left(\boldsymbol{x}_{i}\right)$, $\psi\left(\boldsymbol{x}_{j}\right)$, and $\alpha\left(\boldsymbol{x}_{j}\right)$ are called query, key, and value.
$\delta$ is a position encoding function and $\rho$ is a normalization function such as $\operatorname{softmax}$ (mask block is optional).
The scalar attention layer computes the scalar product between features transformed by $\varphi$ and $\phi$, and uses the output as an attention weight for aggregating features transformed by $\alpha$.

Differently, in the vector attention,  attention weights are vectors that can modulate individual feature channels:
\begin{equation}
\boldsymbol{y}_{i}=\sum_{\boldsymbol{x}_{j} \in \mathcal{X}} \rho\left(\gamma\left(\beta\left(\varphi\left(\boldsymbol{x}_{i}\right), \psi\left(\boldsymbol{x}_{j}\right)\right)+\delta\right)\right) \odot \alpha\left(\boldsymbol{x}_{j}\right)
\end{equation}
where $\beta$ is a relation function (e.g., subtraction) and $\gamma$ is a mapping function (e.g., an MLP) that produces attention vectors for feature aggregation,
$\odot$ is element-wise product.

Nevertheless, scalar attention and vector attention do not perfectly satisfy our requirements.
First, they both employ position encoding to indicate the absolute position of the input token, but we only need a relative relationship.
Second, the value is only a derivative of $\boldsymbol{x}_j$, but we hope it can reflect the relative relationship between $\boldsymbol{x}_{i}$ and $\boldsymbol{x}_{j}$ as well.
Point Transformer~\cite{zhao2021point} proposes a relative position encoding and adds the position encoding to the value item,
alleviating the above two issues.
However, its relative position encoding is additive. The addition represents the translation operation in the vector space, but we need a more flexible operation to manipulate the features from different views.
Moreover, if we directly use Point Transformer in our task, we have to concatenate all the 2D joints and converse it into the position encoding. This procedure results in more parameters.
More parameters but less flexibility often lead to the generalization problem, and this problem is verified in \tablename~\ref{result_generalization}.

\begin{figure}[t]
  \centering
  \includegraphics[width=0.45\textwidth]{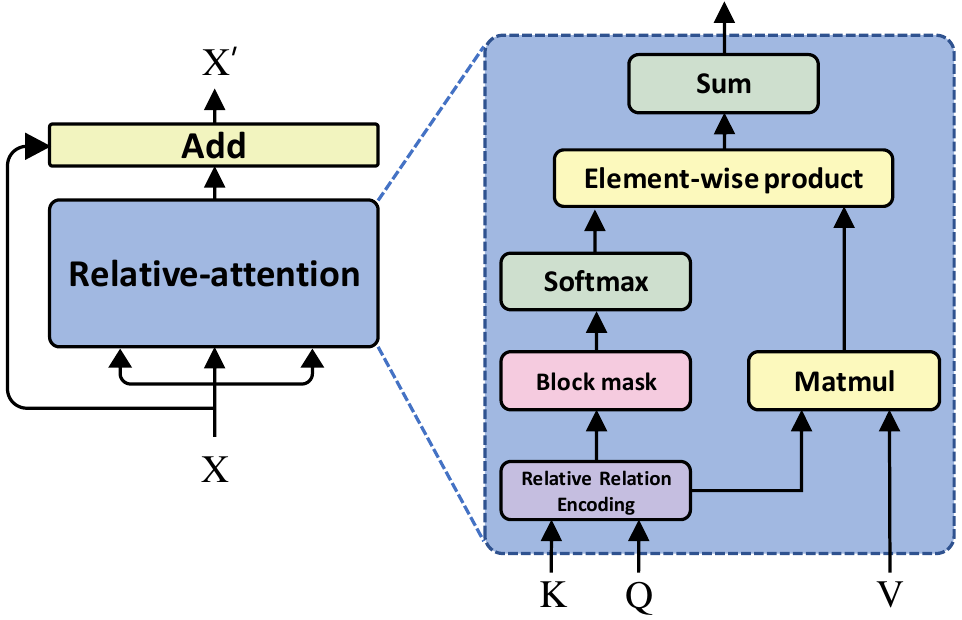}
  \caption{The architecture of Multi-view Fusing Transformer}\label{MTF}
\end{figure}

\subsubsection{Multi-view Fusing Transformer}
To bridge the gap between our purpose and existing transformer models, we propose the Multi-view Fusing Transformer (MFT).
As shown in \figurename~\ref{MTF},
taking $\mathbf{X}\in\mathbb{R}^{C \times N \times T}$ as input, MFT considers it as tokens of $\mathcal{X}=\{\boldsymbol{x}_{i}\}_{i=1}^{N}$, from the perspective of view.
The dimension of $T$ is omitted here as MFT equally operates in each time slice.
In addition, different body parts go through a similar transformation between multiple viewpoints even after feature transformation.
Inspired by Squeeze Reasoning~\cite{9502519} that related components distribute in different groups along channels sparsely,
we divide the dimension of $C$ into $K$ groups, and the same transform matrix manipulates each group.
So, we get $\boldsymbol{x}_i\in \mathbb{R}^{D\times K}$, $C=D\times K$.
The output of MFT is $\mathbf{X}'$:
\begin{equation}
  \mathbf{X}'=\text{RA}\left(\mathbf{X}\right)+\mathbf{X}
\end{equation}
$\text{RA}$ is Relative-Attention.
In Relative-Attention, the input $\mathcal{X}$ triplicates the role of query, key, and value, the output is $\mathcal{Y}=\{\boldsymbol{y_{i}}\}_{i=1}^{N}$.
\begin{align}
  &\mathbf{A}_{ij}=\gamma\left(\Re\left(\boldsymbol{x_i},\boldsymbol{x}_j\right)\right) \\
  &\mathbf{T}_{ij}=\alpha\left(\Re\left(\boldsymbol{x_i},\boldsymbol{x}_j\right)\right) \\
  &\boldsymbol{y}_{i}=\sum_{\mathbf{x}_j\in \mathcal{X}}\rho\left(\mathbf{A}_{ij}\right)\odot\left(\mathbf{T}_{ij}\boldsymbol{x}_j\right)
\end{align}
where
$\Re\left(\boldsymbol{x}_i, \boldsymbol{x}_j\right)$ measures the relationship between each pair of view $\{\boldsymbol{x}_i, \boldsymbol{x}_j\}$,
$\gamma$ and $\alpha$ further transform $\Re\left(\boldsymbol{x}_i, \boldsymbol{x}_j\right)$ into attention matrix $\mathbf{A}_{ij}\in \mathbb{R}^{D\times K}$ and transform matrix $\mathbf{T}_{ij}\in \mathbb{R}^{D\times D}$ via fully connected layers,
$\rho$ consists of a block mask module and a $\operatorname{softmax}$ operation.
The block mask module randomly sets all the values of $\mathbf{A}_{ij}$ to $-inf$ at the rate of $M$, except the condition that $i=j$. Those values turn into zero after $\operatorname{softmax}$.
This mechanism ensures the MFT generalizes well to the scenario with an arbitrary number of views.
For further regularization, we penalize the difference between the inferred $\mathbf{T}_{ij}$ and $\mathbf{T}^{'}_{ij}$ that derived from rotation matrix $\mathbf{R}_{ij}$ between two viewpoints.
$\mathbf{R}_{ij}$ is the rotation matrix calculated from the 3D pose of $i$-th and $j$-th viewpoints via SVD.
Then, it is flattened, transformed with MLP $\psi$, and reshaped as $\mathbf{T}_{ij}$.
\begin{equation}
  \mathbf{T}^{'}_{ij}=\psi(\mathbf{R}_{ij})
\end{equation}
In the testing phase, this branch related to $\mathbf{T}^{'}_{ij}$ is discarded.
It is interesting to note that our framework is also compatible with the scenario with camera parameters.
When utilizing camera extrinsic, we let $\mathbf{T}^{'}_{ij}$ take the place of $\mathbf{T}_{ij}$ in both the training and testing phase.
In this circumstance, $\mathbf{R}_{ij}$ is the rotation matrix of extrinsic parameters.
The architecture of $\Re\left(\boldsymbol{x}_i, \boldsymbol{x}_j\right)$ is shown in \figurename~\ref{relation_encoding}, formulated as:
\begin{figure}[t]
  \centering
  \includegraphics[width=0.49\textwidth]{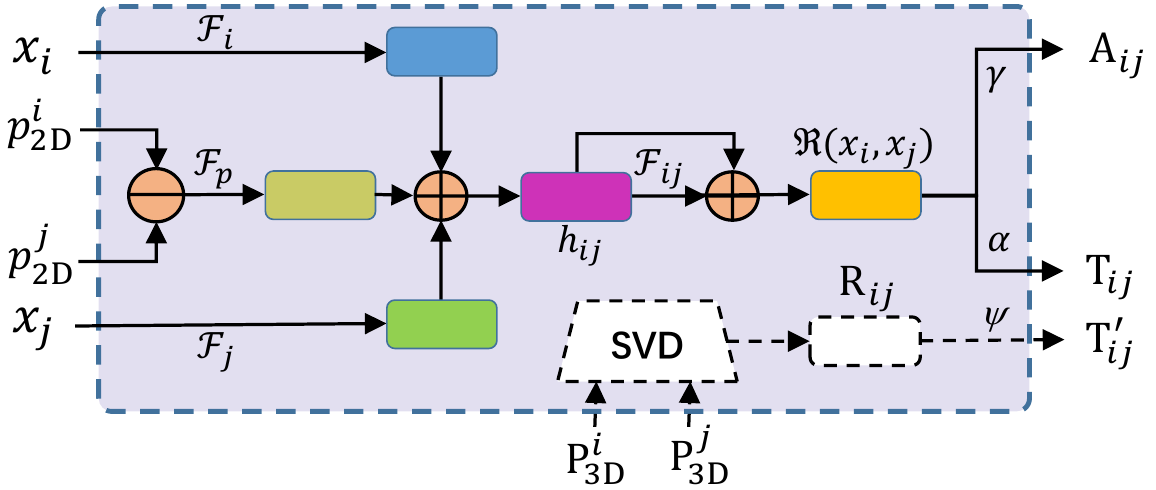}
  \caption{The architecture of relative relation encoding module}\label{relation_encoding}
\end{figure}
\begin{equation}
  \boldsymbol{h}_{ij}=\mathcal{F}_{p}(\boldsymbol{p}_{\text{2D}}^{i}-\boldsymbol{p}_{\text{2D}}^{j})+\mathcal{F}_i\left(\boldsymbol{x}_i\right)+\mathcal{F}_j\left(\boldsymbol{x}_j\right)
\end{equation}
\begin{equation}
  \Re\left(\boldsymbol{x}_i, \boldsymbol{x}_j\right)=\mathcal{F}_{ij}\left(\boldsymbol{h}_{ij}\right)+\boldsymbol{h}_{ij}
\end{equation}
where $\boldsymbol{p}_{\text{2D}}^{i}$ and $\boldsymbol{p}_{\text{2D}}^{j}$ are flatten 2D poses from the 2D detector.
We add the offset between viewpoints to enhance the geometric awareness of the relative-attention module.
$\mathcal{F}_p$, $\mathcal{F}_i$, $\mathcal{F}_j$ and $\mathcal{F}_{ij}$ are fully connected layers.

The Relative-Attention is a vector product-like operation. The difference between them is that (1) the explicit position encoding is discarded in query and value items, and (2) the relative relationship is also integrated into the value item in the form of a transform matrix.
In brief, MFT reconstructs the feature of each view according to the relationship between them, formulated as:
$\mathbf{X}\to\mathbf{X'}, \mathbf{X'}\in\mathbb{R}^{C\times N\times T}$.

\subsection{Temporal Fusing Transformer}
\begin{figure}[h!]
  \centering
  \includegraphics[width=0.49\textwidth]{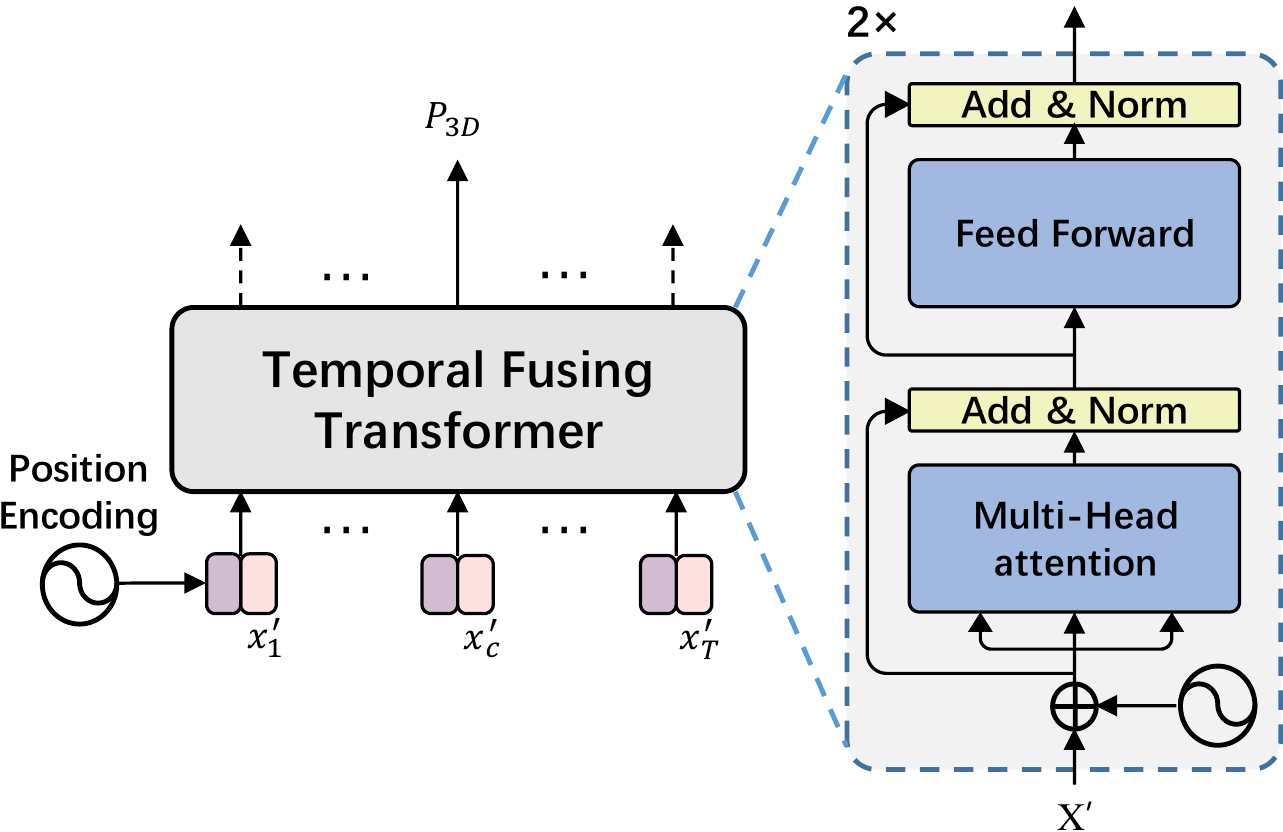}
  \caption{The architecture of Temporal Fusing Transformer. It predicts the 3D pose of the middle frame.}\label{TFT}
\end{figure}

\begin{table*}[h]\normalsize
  \centering
  \caption{Quantitative results on Human3.6M. MTF-Transformer and MTF-Transformer+ are trained with 27 frames, where $T$ is the length of the sequence for testing. We employ CPN~\cite{chen2018cascaded} as the 2D detector, and * means no 2D detector.}
    \resizebox{\textwidth}{!}{
    \begin{tabular}{lc|ccccccccccccccc|c}
    \hline
          & & Dir.  & Disc. & Eat.  & Greet & Phone & Photo & Pose  & Purch. & Sit.  & SitD. & Smoke & Wait  & WalkD. & Walk  & WalkT. & Avg \\
    \hline
    \hline
    \multicolumn{17}{c}{Monocular methods} \\
    \hline
    Pavllo et al.~\cite{pavllo20193d} &(CPN, $T=243$)   & 45.2  & 46.7  & 43.3  & 45.6  & 48.1  & 55.1  & 44.6  & 44.3  & 57.3  & 65.8  & 47.1  & 44.0  & 49.0  & 32.8  & 33.9  & 46.8 \\
    Chen et al.~\cite{li2019generating} &(CPN, $T=1$) & 43.8  & 48.6  & 49.1  & 49.8  & 57.6  & 61.5  & 45.9  & 48.3  & 62.0  & 73.4  & 54.8  & 50.6  & 56.0  & 43.4  & 45.5  & 52.7 \\
    Liu et al.~\cite{liu2020attention} &(CPN, $T=243$)  & 41.8  & 44.8  & 41.1  & 44.9  & 47.4  & 54.1  & 43.4  & 42.2  & 56.2  & 63.6  & 45.3  & 43.5  & 45.3  & 31.3  & 32.2  & 45.1 \\
    Wang et al.~\cite{wang2020motion} &(CPN, $T=96$)   & 40.2  & 42.5  & 42.6  & 41.1  & 46.7  & 56.7  & 41.4  & 42.3  & 56.2  & 60.4  & 46.3  & 42.2  & 46.2  & 31.7  & 31.0  & 44.5 \\
    Zeng et al.~\cite{zeng2020srnet} &(CPN, $T=243$)    & 46.6  & 47.1  & 43.9  & 41.6  & 45.8  & 49.6  & 46.5  & 40.0  & 53.4  & 61.1  & 46.1  & 42.6  & 43.1  & 31.5  & 32.6  & 44.8 \\
    Cheng et al.~\cite{cheng2019occlusion} &(CPN, $T=128$) & 38.3 &41.3 &46.1 &40.1 &41.6 &51.9 &41.8 &40.9 &51.5 &58.4 &42.2 &44.6 &41.7 &33.7 &30.1 &42.9 \\
    \hline
    \hline
    \multicolumn{17}{c}{Multi-view methods with camera parameters} \\
    \hline
    Pavlakos et al.~\cite{pavlakos2017harvesting} &(*, $T=1$) & 41.2  & 49.2  & 42.8  & 43.4  & 55.6  & 46.9  & 40.3  & 63.7  & 97.6  & 119   & 52.1  & 42.7  & 51.9  & 41.8  & 39.4  & 56.9 \\
    Qiu et al.~\cite{qiu2019cross} &(*, $T=1$) & 24.0  & 26.7  & 23.2  & 24.3  & 24.8  & 22.8  & 24.1  & 28.6  & 32.1  & 26.9  & 31.0  & 25.6  & 25.0  & 28.0  & 24.4  & 26.2 \\
    Iskakov et al.~\cite{iskakov2019learnable} &(*, $T=1$) & 19.9  & 20.0  & 18.9  & 18.5  & 20.5  & 19.4  & 18.4  & 22.1  & 22.5  & 28.7  & 21.2  & 20.8  & 19.7  & 22.1  & 20.2  & 20.8 \\
    He et al.(IMU)~\cite{he2020epipolar} &(*, $T=1$) & 25.7  & 27.7  & 23.7  & 24.8  & 26.9  & 31.4  & 24.9  & 26.5  & 28.8  & 31.7  & 28.2  & 26.4  & 23.6  & 28.3  & 23.5  & 26.9 \\
    Zhang et al.~\cite{zhang2021adafuse} &(*, $T=1$) & 17.8  & 19.5  & 17.6  & 20.7  & 19.3  & 16.8  & 18.9  & 20.2  & 25.7  & 20.1  & 19.2  & 20.5  & 17.2  & 20.5  & 17.3  & 19.5 \\
    Zhang et al.~\cite{zhang2020fusing} &(*, $T=1$)          & —     & —     & —     & —     & —     & —     & —     & —     & —     & —     & —     & —     & —     & —     & —     & 21.7 \\
    Remeli et al.~\cite{remelli2020lightweight} &(*, $T=1$)   & 27.3  & 32.1  & 25.0  & 26.5  & 29.3  & 35.4  & 28.8  & 31.6  & 36.4  & 31.7  & 31.2  & 29.9  & 26.9  & 33.7  & 30.4  & 30.2 \\
    MTF-Transformer+ &(CPN, $T=1$)                  & 23.8  & 26.0  & 23.9  & 25.0  & 28.2  & 29.7  & 23.6  & 25.5  & 30.1  & 37.3  & 26.6  & 24.5  & 27.4  & 23.1  & 23.4  & 26.5
    \\
    MTF-Transformer+ &(CPN, $T=27$)                 & 23.4  & 25.2  & 23.1  & 24.4  &	27.4  & 28.5  & 22.8  & 25.2  & 28.7  & 36.2  & 25.9  & 23.6  & 26.6  & 22.6  & 22.7  & 25.8
    \\
    \hline
    \hline
    \multicolumn{17}{c}{Multi-view methods without camera parameters} \\
    \hline
    Huang et al.~\cite{huang2020deepfuse} & ($*, T=1$) & 26.8  & 32.0  & 25.6  & 52.1  & 33.3  & 42.3  & 25.8  & 25.9  & 40.5  & 76.6  & 39.1  & 54.5  & 35.9  & 25.1  & 24.2  & 37.5 \\
    FLEX~\cite{gordon2021flex} & (\cite{iskakov2019learnable}, $T=27$)
                                          & \textbf{23.1}  & 28.8  & 26.8  & 28.1  & 31.6  & 37.1  & 25.7  & 31.4  & 36.5  & 39.6  & 35.0  & 29.5  & 35.6  & 26.8  & 26.4  & 30.9 \\
    FLEX~\cite{gordon2021flex}&(CPN, $T=27$) & —     & —     & —     & —     & —     & —     & —     & —     & —     & —     & —     & —     & —     & —     & — & 31.7 \\
    MTF-Transformer &(CPN, $T=1$)          & 24.2  & 26.4  & 26.1  & 25.6  & 29.4  & 29.7  & 25.1  & 25.4  & 32.4  & 37.4  & 27.1  & 25.4  & 29.5  & 23.8  & 24.4  & 27.5 \\
    MTF-Transformer &(CPN, $T=27$)         & \textbf{23.1}  & \textbf{25.4}  & \textbf{24.7}  & \textbf{24.5}  & \textbf{27.9}  & \textbf{28.3}  & \textbf{23.9}  & \textbf{24.6}  & \textbf{30.7}  & \textbf{35.7}  & \textbf{25.8}  & \textbf{24.2}  & \textbf{28.4}  & \textbf{22.8}  & \textbf{23.1}  & \textbf{26.2} \\
    MTF-Transformer &\begin{tabular}[c]{@{}c@{}}(CPN, $T=27$)\\ no added view \end{tabular} & 24.6  & \textbf{25.4}  & 24.8  & 24.6  & 28.7  & 29.1  & \textbf{23.9}  & 25.6  & 31.4  & 36.2  & 26.6  & 24.7  & 28.9  & 23.7  & 23.6  & 26.6 \\
    \hline
    FLEX~\cite{gordon2021flex}&(GT, $T=27$) & —     & —     & —     & —     & —     & —     & —     & —     & —     & —     & —     & —     & —     & —     & —     & 22.9 \\
    MTF-Transformer &(GT, $T=27$)~         & 15.5  & 17.1  & 13.7  & 15.5  & 14.0  & 16.2  & 15.8  & 16.5  & 15.8  & 16.1  & 14.5  & 14.5  & 16.9  & 14.3  & 13.7  & 15.3 \\
    \hline
    \end{tabular}%
  \label{result_human}%
  }
\end{table*}%

The Temporal Fusing Transformer (TFT) is shown in \figurename~\ref{TFT}, it takes $\mathbf{X'}$ as input and predicts the 3D pose of $J$ joint points $P_{3D}\in \mathbb{R}^{3\times J \times N}$ in static scenes or dynamic scenes.
Specifically, TFT utilizes a Transformer Encoder block~\cite{vaswani2017attention} of two encoder layers to get the 3D pose of the middle frame.
As the temporal sequence has a direction and the order of frames matters, the position encoding is employed here.
In addition, TFT masks some frames during the training stage to be compatible with a single image in static scenes and multi-view videos in dynamic scenes.
For example, when the input video sequence has 7 frames, the left and right frames are masked evenly.

\subsection{Loss Function}
The loss Function consists of two components.
We employ the mean per joint position error (MPJPE) as the training loss and the testing metric.
MPJPE first aligns the root joint (central hip) of predicted skeleton $\mathcal{S}=\{p_i\}_{i=1}^{J}$ and the ground truth skeleton $\mathcal{S}^{gt}=\{p^{gt}_i\}_{i=1}^{J}$,
and then calculates the average Euclidean distance between each joints of them.
MPJPE is computed as:
\begin{equation}
\label{MPJPE}
L_{M}(\mathcal{S})=\frac{1}{J} \sum_{i=1}^{J} \| p_{i}-p^{gt}_{i} \|_{2}
\end{equation}
Besides, we utilize the rotation matrix between each pair of views to constrain the transform matrix $\mathbf{T}_{ij}$, an extra transform error is also used as:
\begin{equation}
L_{t}=\|\mathbf{T}_{ij} - \mathbf{T}^{'}_{ij}\|_{1}
\end{equation}
The total loss function is:
\begin{equation}
  L=L_{M}+\lambda L_{t}
\end{equation}

\subsection{Implementation Details}
MTF-Transformer is an end-to-end method implemented with Pytorch.
We employ a pretrained 2D detector with frozen weights in the training stage.
During the training phase, batch size, learning rate, learning decay, and dropout rate are set to $720$, $1e^{-3}$, 0.95, 0.1, respectively.
Note that learning decay is executed after the end of every epoch.
We adopt the same strategy for BN momentum decay as in \cite{pavllo20193d} and use Adam Optimizer for all modules.
Besides, we set the channel $C$ to 600 and the $\lambda$ in the loss function to 0.5 and train the model with 60 epochs.

\section{Experiments}
In this section, we first report quantitative and qualitative results of MTF-Transformer on three datasets.
Then, we conduct ablation studies to verify the effectiveness of our design in all modules.
Considering the clarity and brevity of this section, we place some ablation studies on the hyper-parameters in the appendix part.
\begin{table*}[htbp]
  \centering
  \renewcommand{\arraystretch}{1.1}
  \caption{Quantitative results on TotalCapture. We employ HRNet-W32(D1), ResNet50(D2), ResNet101(D3) as 2D pose detector. MTF-Transformer and MTF-Transformer+ are trained with 27 frames, where $T$ is the length of the sequence for testing.}
  \resizebox{\textwidth}{!}{
    \begin{tabular}{l|ccc|ccc|c|ccc|ccc|c}
    \hline
    \multicolumn{1}{c|}{\multirow{3}{*}{Methods}} & \multicolumn{7}{c|}{Seen Cameras(1,3,5,7)}      & \multicolumn{7}{c}{Unseen Cameras(2,4,6,8)} \\
    \cline{2-15}
    & \multicolumn{3}{c|}{Seen Subjects(S1, S2, S3)} & \multicolumn{3}{c|}{Unseen Subjects(S4, S5)} & \multirow{2}{*}{Mean}
    & \multicolumn{3}{c|}{Subjects(S1, S2)} & \multicolumn{3}{c|}{Unseen Subjects(S4, S5)} & \multirow{2}{*}{Mean} \\
    \cline{2-7}\cline{9-14}
                         & W2    & FS3   & A3    & W2    & FS3   & A3    &       & W2    & FS3   & A3    & W2    & FS3   & A3    &      \\
    \hline
    \hline
    \multicolumn{15}{c}{Multi-view methods with camera parameters} \\
    \hline
    Qiu et al.~\cite{qiu2019cross}                     & 19.0  & 28.0  & 21.0  & 32.0  & 54.0  & \textbf{33.0}  & 29.0  & ——    & ——    & ——    & ——    & ——    & ——    & ——   \\
    Remeli et al.~\cite{remelli2020lightweight}        & \textbf{10.6}  & 30.4  & \textbf{16.3}  & \textbf{27.0}  & 65.0  & 34.2  & 27.5  & 22.4  & 47.1  & 27.8  & 39.1  & 75.7  & 43.1  & 38.2 \\
    MTF-Transformer+(D1,$T=1$)                           & 12.3  & 28.3  & 18.5  & 27.2  & 49.1  & 33.4  & 26.0  & 16.2  & 31.4  & 20.2  & 29.3  & 51.1  & 36.3  & 28.6 \\
    MTF-Transformer+(D1,$T=27$)                          & 11.5  & 27.4  & 17.8  & \textbf{27.0}  & \textbf{48.5}  & 33.3  & 25.4  & 15.6  & 30.5  & 19.4  & \textbf{29.2}  & 50.6  & 36.2  & 28.1 \\
    MTF-Transformer+(D2,$T=1$)                           & 11.8  & 28.3  & 18.0  & 27.4  & 50.2  & 33.7  & 26.0  & 15.6  & 31.3  & 19.7  & 29.7  & 53.3  & \textbf{35.0}  & 28.5 \\
    MTF-Transformer+(D2,$T=27$)                          & 11.2  & 27.5  & 17.3  & 27.4  & 49.3  & 33.7  & 25.4  & 15.2  & 30.7  & 18.9  & 29.6  & 52.5  & 35.2  & 28.1 \\
    MTF-Transformer+(D3,$T=1$)                           & 11.4  & 27.5  & 17.3  & 27.5  & 50.2  & 34.1  & 25.7  & 14.5  & 30.0  & 18.8  & 29.4  & 50.1  & 35.5  & 27.5 \\
    MTF-Transformer+(D3,$T=27$)                          & 10.7  & \textbf{26.5}  & 16.7  & 27.4  & 49.4  & 34.1  & \textbf{25.1}  & \textbf{13.9}  & \textbf{29.2}  & \textbf{18.1}  & \textbf{29.2}  & \textbf{49.5}  & 35.6  & \textbf{27.0} \\
    \hline
    \hline
    \multicolumn{15}{c}{Multi-view methods without camera parameters} \\
    \hline
    FLEX~\cite{gordon2021flex} (D1,T=27)     & 38.3  & 80.8  & 39.7  & 40.0  & 131.2 & 57.7  & 50.2  & 107.3 & 149.6 & 103.1 & 116.7 & 190.2 & 106.8 & 120.2 \\
    FLEX~\cite{gordon2021flex} (D2,T=27)     & 34.5  & 78.3  & 36.4  & 39.3  & 128.3 & 59.4  & 48.2  & 106.4 & 141.7 & 103.7 & 114.8 & 177.3 & 122.3 & 119.7 \\
    FLEX~\cite{gordon2021flex} (D3,T=27)     & 33.2  & 81.0  & 34.2  & 38.3  & 123.8 & 59.5  & 49.4  & 109.3 & 152.1 & 105.3 & 114.3 & 175.5 & 122.5 & 125.4 \\
    MTF-Transformer(D1,$T=1$)     & 11.1  & 30.0  & 16.3  & 26.0  & 53.4  & 32.9  & 25.9  & 26.3  & 44.7  & 30.2  & 37.6  & 66.3  & 43.9  & 39.3 \\
    MTF-Transformer(D1,$T=27$)    & 9.8   & 27.8  & 14.9  & \textbf{25.8}  & \textbf{51.6}  & \textbf{32.7}  & \textbf{24.6}  & 25.7  & 43.4  & 29.4  & 37.4  & 64.7  & 44.1  & 38.6 \\
    MTF-Transformer(D2,$T=1$)     & 10.9  & 29.8  & 16.2  & 26.9  & 54.2  & 33.5  & 26.1  & 26.3  & 45.3  & 29.9  & 38.4  & 66.5  & 43.4  & 39.5 \\
    MTF-Transformer(D2,$T=27$)    & 9.7   & 27.8  & 14.9  & 26.6  & 52.4  & 33.3  & 24.9  & 25.8  & 44.0  & 29.2  & 38.2  & 65.0  & 43.6  & 38.8 \\
    MTF-Transformer(D3,$T=1$)     & 10.5  & 28.4  & 15.6  & 26.9  & 54.7  & 33.8  & 25.7  & 24.2  & 41.4  & 28.1  & 37.1  & 63.0  & \textbf{42.4}  & 37.2 \\
    MTF-Transformer(D3,$T=27$)    & \textbf{9.3 }  & \textbf{26.5}  & \textbf{14.5}  & 26.7  & 53.1  & 33.8  & 24.7  & \textbf{23.7}  & \textbf{40.3}  & \textbf{27.4}  & \textbf{37.0}  & \textbf{61.8}  & 42.9  & \textbf{36.6} \\
    \hline
    \end{tabular}%
  \label{result_totalcapture}%
  }
\end{table*}%
\subsection{Datasets}
We evaluate MTF-Transformer on three datasets, including:\\
\textbf{Human3.6M (H36M)}~\cite{ionescu2013human3} is a large publicly available 3D human pose benchmark for both monocular and multi-view setups.
It consists of 3.6 million image frames from 4 synchronized 50Hz digital cameras, and the corresponding 2D pose and 3D pose are captured by the MoCap system in a constrained indoor studio environment.
Each actor performs 15 everyday activities such as walking, discussing, etc.
Following previous works~\cite{lee2018propagating,sun2017compositional, martinez2017simple}, we use 5 subjects (S1, S5, S6, S7, S8) for training and 2 subjects (S9, S11) for testing, and report MPJPE~\cite{2019Trajectory, pavllo20193d,cai2019exploiting} as the evaluation metric.
We simulate an additional virtual view when training MTF-Transformer to enhance its flexibility.
The 2D pose in the virtual view is synthesized via random rotation and projection, following Cheng et al.~\cite{cheng2019occlusion}.
To verify the effectiveness of the virtual view, we also report the result of MTF-Transformer trained with no added view.\\
\textbf{TotalCapture}~\cite{joo2018total} is captured from 8 calibrated full HD video cameras recording at 60Hz. It features five subjects. Each subject performs four diverse performances 3 times, involving ROM, Walking, Acting, and Freestyle. Accurate 3D human joint locations are obtained from a marker-based motion capture system. Following previous work, the training set consists of “ROM1,2,3”,

“Walking1,3”, “Freestyle1,2”, “Acting1,2”, on subjects 1,2, and 3. The test set consists of “Walking2 (W2)”, “Freestyle3 (FS3)”, and “Acting3 (A3)” on subjects 1, 2, 3, 4, and 5. The number following each action indicates the video from which the action is. For example, Freestyle has three videos of the same action, of which 1 and 2 are used for training and 3 for testing.
Camera 1,3,5,7 is used in both the training and testing set, but camera 2,4,6,8 only appear in the testing set.
That is to say. The testing set has some unseen camera configuration.\\
\textbf{KTH Multiview Football II}~\cite{kazemi2013multi} consists of 8000+ images of professional footballers during a match in the Allsvenskan league. It is filmed by moving cameras and contains 14 joints(top-head, neck, shoulders, hips, knees, feet, elbows, and hands). To match the topology of H36M, we create the root (pelvis) by averaging the hips, the nose by averaging the neck and top-head, and the spine by averaging the root and the neck.

\begin{figure*}[ht]
	\centerline{\includegraphics[width=1.0\textwidth]{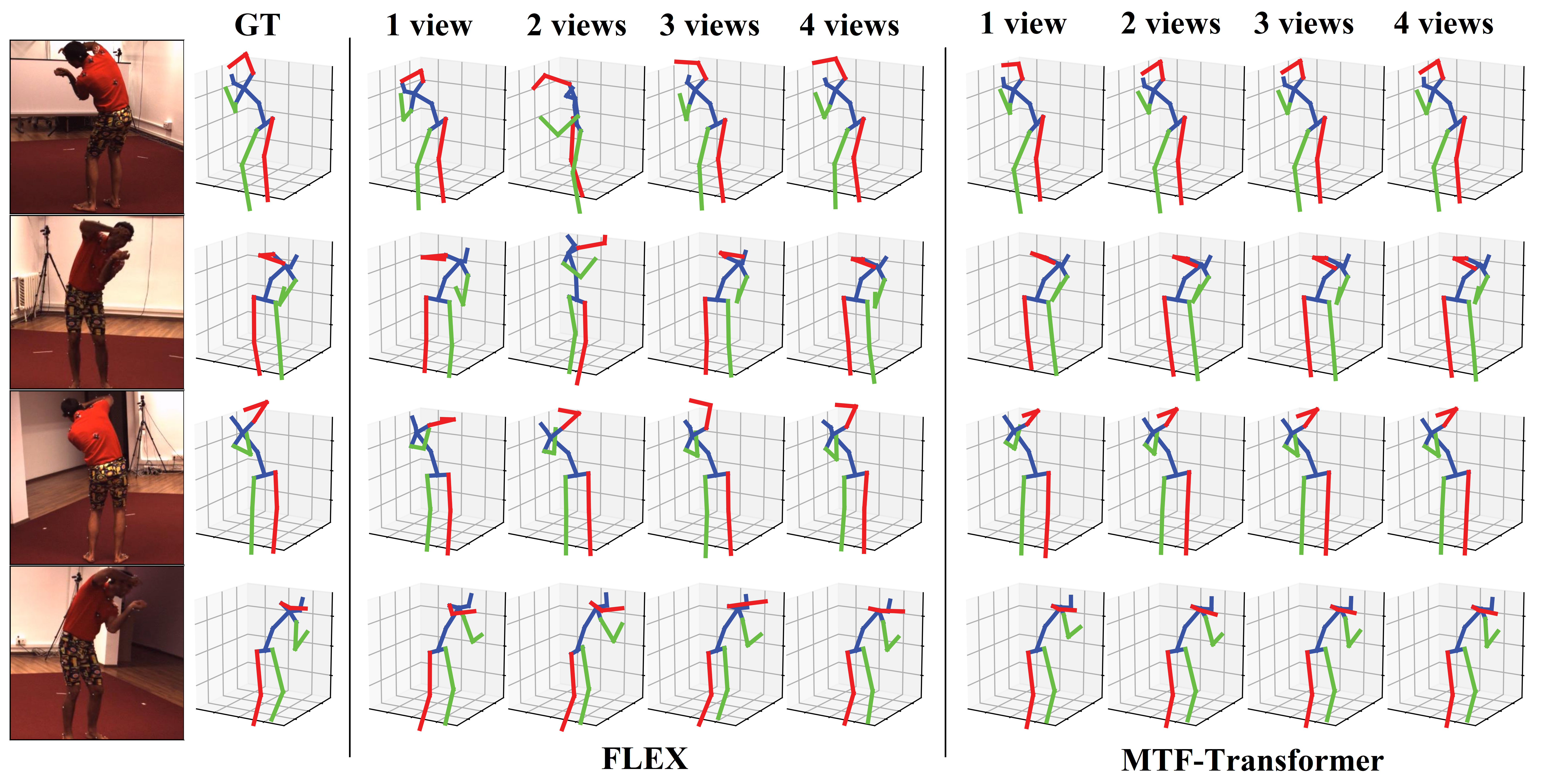}}
	\caption{Results of FLEX and MTF-Transformer with different view numbers on the Human3.6M}
	\label{fig_vis_result_h36m}
\end{figure*}

\begin{figure*}[h]
	\centerline{\includegraphics[width=0.7\textwidth]{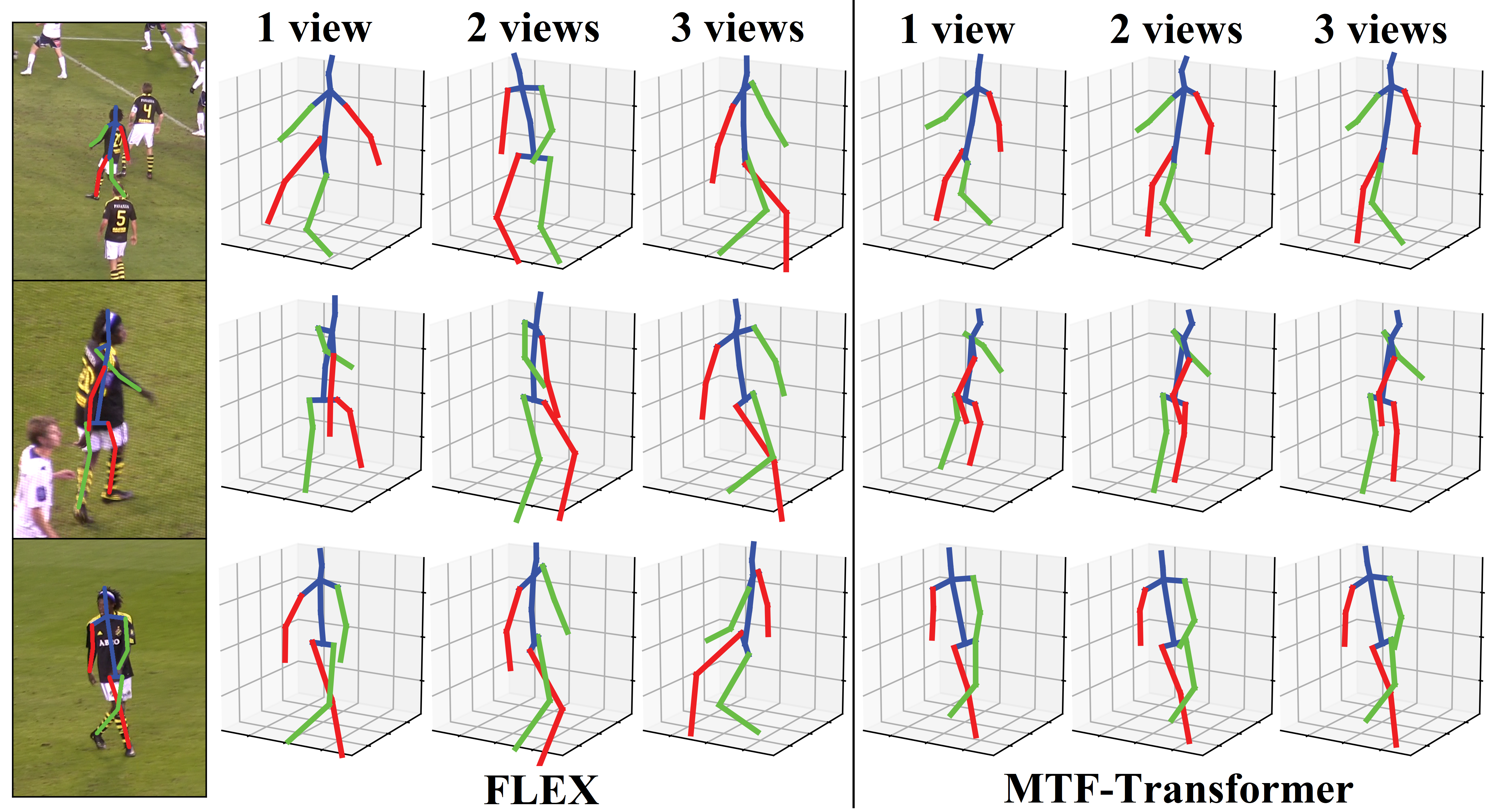}}
	\caption{Demonstration of transfer FLEX and MTF-Transformer trained on Human3.6M to KTH Multiview Football II}
	\label{fig_vis_result}
\end{figure*}

\subsection{Quantitative Evaluation}
We report the quantitative results of MTF-Transformer on Human3.6M and TotalCapture. MTF-Transformer and MTF-Transformer+ represent the vanilla MTF-Transformer and the MTF-Transformer utilizing camera parameters, respectively. In MTF-Transformer+, we directly use the transform matrix $\mathbf{T}^{'}_{ij}$ calculated from rotate matrix $\mathbf{R}_{ij}$ between 3D ground truth.
\\
\textbf{Human3.6M}:
The quantitative results of MTF-Transformer and competitive methods are shown in \tablename~\ref{result_human}.
When CPN is used as a 2D pose detector,
MTF-Transformer outperforms all the monocular methods, and it decreases the MPJPE by 1.3 when increasing the length of sequence from 1 to 27,
indicating that multi-view and temporal information benefits for 3D pose estimation.
When we employ Ground Truth as 2D pose input, both \cite{gordon2021flex} and MTF-Transformer obtain significant improvement,
indicating that 2D pose plays an essential role in 2D-to-3D methods.
Compared to multi-view methods with camera calibration, MTF-Transformer is superior to~\cite{pavlakos2017harvesting}, ~\cite{remelli2020lightweight}, and ~\cite{he2020epipolar} but inferior to others.
It shows that MTF-Transformer is competitive, but camera calibration still has an obvious advantage.
Compared to Multi-view methods without calibration, MTF-Transformer achieves the best performance and demonstrates its superiority.
Considering the difficulty of calibrating the camera in real-time, MTF-Transformer is a satisfactory attempt.
Besides, when we extend to MTF-Transformer+, we further improve the result.
MTF-Transformer+ is inferior to some calibration-need methods.
Considering that our focus is on fusing multi-view features without calibration and those superior methods utilize extra sensors or 2D image features,
the performance of the MTF-Transformer+ is acceptable.
\\
\textbf{TotalCapture}: The quantitative results of MTF-Transformer and competitive methods are shown in \tablename~\ref{result_totalcapture}.
MTF-Transformer series are trained on camera 1, 3, 5, 7 of the training set, and tested on camera 1, 3, 5, 7 (seen) and camera 2, 4, 6, 8 (unseen) of the testing set.
The testing set includes both seen subjects and unseen subjects in the training set.
From the vertical comparison, MTF-Transformer+ outperforms ~\cite{qiu2019cross} and~\cite{remelli2020lightweight} with arbitrary 2D detector and the length of sequence,
and MTF-Transformer has superior result over ~\cite{gordon2021flex}.
Besides, the 2D detector has an influence on the result, and increasing the length of the sequence improves the performance.
Moreover, MTF-Transformer with ResNet101 as 2D detector obtains better result than ~\cite{qiu2019cross} and~\cite{remelli2020lightweight} demonstrating the superiority of our method.
From horizontal analysis, all the methods achieve better performance on seen cameras than on unseen cameras, on seen subjects than on unseen subjects.
It means that generalization is an important issue for 3D pose estimation.

\subsection{Qualitative Evaluation}
Some results of FLEX and MTF-Transformer on Human3.6M are shown in \figurename~\ref{fig_vis_result_h36m}.
Both FLEX and MTF-Transformer improve the prediction as the number of views increases, but MTF-Transformer has better results when the number of views is low.
The reason is that MTF-Transformer uses CAA to reduce the influence of 2D detector errors and introduces Relative-Attention to improve the fusion efficiency between multi-view features.
To further verify the generalization of MTF-Transformer under different camera configurations, we test the model trained on Human3.6M on more challenging KTH Multiview Football II. Some results of generalization experiments are shown in Fig.~\ref{fig_vis_result}.
It demonstrates that MTF-Transformer can generalize well from an indoor lab scene to the wild environment because it stands free from camera parameters and measures the implicit relationship between views adaptively.
Although FLEX is also parameter-free, it aggregates the features from multiple viewpoints into one feature and then splits it into different viewpoints.
The viewpoint awareness is twisted in the procedure by FLEX while MTF-Transformer keeps each viewpoint's independence.

\subsection{Ablation Study}
In this section, we verify the effectiveness of all modules of MTF-Transformer on Human3.6M.
We train all the models with 5 views (4 cameras and an additional synthesized view) and test them with different views unless otherwise stated.
To eliminate the effect of the 2D detector, we take 2D detection from CPN~\cite{chen2018cascaded} as input.

\subsubsection{Analysis on Confidence Attentive Aggregation}
\begin{table}[htbp]
  \centering
  \renewcommand{\arraystretch}{1.1}
  \caption{Results of different procedures to fuse the 2D pose and the confidence from 2D detector on Human3.6M.}
    \begin{tabular}{c|c|c|c|c|c|c}
    \hline
    \multirow{2}{*}{} & \multirow{2}{*}{\begin{tabular}[c]{@{}c@{}}Sequence \\ length $T$\end{tabular}}
    & \multicolumn{4}{c|}{Number of Views N} & \multirow{2}{*}{\begin{tabular}[c]{@{}c@{}}Parameters\\(M)\end{tabular}} \\
    \cline{3-6}
                                                           &       & 1     & 2     & 3     & 4      &                                \\
    \hline
    no confidence     & \multirow{3}{*}{1}                 & 52.3   & 36.8  & 31.6  & 29.4   &   9.8 \\
    \cline{1-1}\cline{3-7}
    concatenate       &                                    & 52.2   & 36.7  & 31.4  & 29.2   &   9.9 \\
    \cline{1-1}\cline{3-7}
    CAA               &                                    & \textbf{50.7}   & \textbf{35.3}  & \textbf{30.1}  & \textbf{28.0}   &  10.1 \\
    \hline
    no confidence     & \multirow{3}{*}{3}                 & 51.3   & 35.9  & 30.8  & 28.7   &   9.8 \\
    \cline{1-1}\cline{3-7}
    concatenate       &                                    & 51.3   & 35.9  & 30.7  & 28.6   &   9.9 \\
    \cline{1-1}\cline{3-7}
    CAA               &                                    & \textbf{49.8}   & \textbf{34.5}  & \textbf{29.4}  & \textbf{27.3}   &  10.1 \\
    \hline
    no confidence     & \multirow{3}{*}{5}                 & 51.0   & 35.6  & 30.6  & 28.5   &   9.8 \\
    \cline{1-1}\cline{3-7}
    concatenate       &                                    & 50.9   & 35.6  & 30.4  & 28.3   &   9.9 \\
    \cline{1-1}\cline{3-7}
    CAA               &                                    & \textbf{49.4}   & \textbf{34.2}  & \textbf{29.2}  & \textbf{27.1}   &  10.1 \\
    \hline
    no confidence     & \multirow{3}{*}{7}                 & 50.7   & 35.4  & 30.5  & 28.4   &   9.8 \\
    \cline{1-1}\cline{3-7}
    concatenate       &                                    & 50.8   & 35.4  & 30.3  & 28.2   &   9.9 \\
    \cline{1-1}\cline{3-7}
    CAA               &                                    & \textbf{49.2}   & \textbf{34.1}  & \textbf{29.1}  & \textbf{27.1}   &  10.1 \\
    \hline
    \end{tabular}%
  \label{result_conf}%
\end{table}%

\begin{table}[h]\normalsize
  \centering
  \renewcommand{\arraystretch}{1.2}
  \caption{The results of different design of CAA}\label{CAA}
  \begin{tabular}{c|c}
    \hline
    Method & MPJPE  \\
    \hline
    $\boldsymbol{f}^{g}=\boldsymbol{\bar{f}}^{g}+\mathbf{a}^{g}\cdot\boldsymbol{p}_\text{2D}^{g}$   & 27.121 \\
    $\boldsymbol{f}^{g}=\mathcal{F}_{res}^{g}\left(\mathbf{a}^{g}\cdot\boldsymbol{p}_\text{2D}^{g}\right)$   & 27.499 \\
    $\boldsymbol{f}^{g}=\mathcal{F}_{res}^{g}\left(\boldsymbol{\bar{f}}^{g}+\mathbf{a}^{g}\cdot\boldsymbol{\bar{f}}^{g}\right)$   & 27.121 \\
    $\boldsymbol{f}^{g}=\mathcal{F}_{res}^{g}\left(\boldsymbol{\bar{f}}^{g}+\mathbf{a}^{g}\cdot\boldsymbol{p}_\text{2D}^{g}\right)$   & 27.056 \\
    \hline
  \end{tabular}
\end{table}

MTF-Transformer employs the Confidence Attentive Aggregation (CAA) module in Feature Extractor to reduce the impact of the unreliable 2D pose.
We report the results of MTF-Transformer with and without CAA. Besides, we also evaluate the technique of concatenating the 2D pose and confidence values.
As shown in \tablename~\ref{result_conf}, concatenating can improve the performance, compared with the circumstance without confidence.
When CAA takes the place of concatenating, MTF-Transformer can achieve better performance at all the number of views.

We also conduct experiments to verify the design of CAA.
When we remove the res-blocks, the result of CAA is $\boldsymbol{f}^{g}=\boldsymbol{\bar{f}}^{g}+\mathbf{a}^{g}\cdot\boldsymbol{p}_\text{2D}^{g}$;
When we remove the shortcut connection, the result of CAA is $\boldsymbol{f}^{g}=\mathcal{F}_{res}^{g}\left(\mathbf{a}^{g}\cdot\boldsymbol{p}_\text{2D}^{g}\right)$;
If we let $\boldsymbol{\bar{f}}^{g}$ takes the place of $\boldsymbol{p}_\text{2D}^{g}$, the result of CAA is $\boldsymbol{f}^{g}=\mathcal{F}_{res}^{g}\left(\boldsymbol{\bar{f}}^{g}+\mathbf{a}^{g}\cdot\boldsymbol{\bar{f}}^{g}\right)$.
As shown in \tablename~\ref{CAA}, when we modulate the 2D pose and employ both shortcut and res-blocks, CAA achieves the best performance.

\subsubsection{Analysis on Multi-view Fusing Transformer}
The Multi-view Fusing Transformer (MFT) measures the relationship between each pair of views and reconstructs the features according to the relationship.
To validate the effectiveness of MFT, we compare its result with other multi-view fusing methods on Human3.6M,
in the aspects of precision and generalization capability.
\begin{table}[ht!]
  \centering
  \renewcommand{\arraystretch}{1.1}
  \caption{Results of different relative attention modules on Human3.6M. $T$ is the length of sequence}
    \begin{tabular}{c|c|c|c|c|c|c}
    \hline
    \multirow{2}{*}{Method} &
    \multirow{2}{*}{$T$} & \multicolumn{4}{c|}{Number of Views N}
    & \multirow{2}{*}{\begin{tabular}[c]{@{}c@{}}Parameters\\(M)\end{tabular}} \\
    \cline{3-6}          &       & 1     & 2     & 3     & 4     &  \\
    \hline
    transformer & \multirow{4}{*}{1} & 50.8  & 38.7 & 34.0 & 31.9 & 10.4 \\
    \cline{1-1}\cline{3-7}    point transformer &       & 51.2  & 36.1 & 30.8 & 28.4 & 11.7 \\
    \cline{1-1}\cline{3-7}    MFT w/o $\mathbf{T}_{ij}$ &       & \textbf{50.6}  & 37.8 & 32.8 & 30.5 & 9.7 \\
    \cline{1-1}\cline{3-7}    MFT  &       & 50.7  & \textbf{35.3} & \textbf{30.1} & \textbf{28.0} & 10.1 \\
    \hline
    transformer & \multirow{4}{*}{3} & 49.9  & 37.8  & 33.1  & 31.0  & 10.4 \\
    \cline{1-1}\cline{3-7}    point transformer &       & 50.2  & 35.2  & 30.1  & 27.9  & 11.7 \\
    \cline{1-1}\cline{3-7}    MFT w/o $\mathbf{T}_{ij}$ &       & \textbf{49.8}  & 37.0  & 32.0  & 29.7  & 9.7 \\
    \cline{1-1}\cline{3-7}    MFT  &       & \textbf{49.8}  & \textbf{34.5}  & \textbf{29.4}  & \textbf{27.3}  & 10.1 \\
    \hline
    transformer & \multirow{4}{*}{5} & 49.6  & 37.5  & 32.8  & 30.7  & 10.4 \\
    \cline{1-1}\cline{3-7}    point transformer &       & 49.9  & 35.0  & 29.9  & 27.7  & 11.7 \\
    \cline{1-1}\cline{3-7}    MFT w/o $\mathbf{T}_{ij}$ &       & 49.5  & 36.6  & 31.6  & 29.4  & 9.7 \\
    \cline{1-1}\cline{3-7}    MFT  &       & \textbf{49.4}  & \textbf{34.2}  & \textbf{29.2}  & \textbf{27.1}  & 10.1 \\
    \hline
    transformer & \multirow{4}{*}{7} & 49.4  & 37.3  & 32.6  & 30.6  & 10.4 \\
    \cline{1-1}\cline{3-7}    point transformer &       & 49.7  & 34.8  & 29.8  & 27.6  & 11.7 \\
    \cline{1-1}\cline{3-7}    MFT w/o $\mathbf{T}_{ij}$ &       & 49.3  & 36.4  & 31.5  & 29.3  & 9.7 \\
    \cline{1-1}\cline{3-7}    MFT  &       & \textbf{49.2}  & \textbf{34.1}  & \textbf{29.1}  & \textbf{27.1}  & 10.1 \\
    \hline
    \end{tabular}%
  \label{relative_attention}%
\end{table}%

\begin{table}[ht!]
  \centering
  \renewcommand{\arraystretch}{1.1}
  \caption{Generalization capability of different relative attention modules. We train all the models on Human3.6M with 2 views, test them with different number of views.}\label{result_generalization}
  \begin{tabular}{c|c|c|c|c|c|c}
    \hline
    \multirow{2}{*}{Method} & \multirow{2}{*}{$T$}
    & \multicolumn{4}{c|}{Number of views $N$} & \multirow{2}{*}{\begin{tabular}[c]{@{}c@{}}Parameters \\ (M)\end{tabular}} \\
    \cline{3-6}
                            &                                      & 1 & 2 & 3 & 4                           &  \\
    \hline
    transformer             & \multirow{4}{*}{1}                   & \textbf{55.9}    & 47.1   & 43.6   & 41.2   & 10.4 \\
    \cline{1-1}\cline{3-7}
    point transformer       &                                      & 58.0    & 51.8   & 49.9   & 48.7   & 11.7 \\
    \cline{1-1}\cline{3-7}
    no MFT                  &                                       & \multicolumn{4}{c|}{57.0}          & 7.5  \\
    \cline{1-1}\cline{3-7}
    MFT         &                                      & 56.4    & \textbf{46.1}   & \textbf{41.5}   & \textbf{39.0}   & 10.1 \\
    \cline{1-1}\cline{3-7}
    \hline
    transformer             & \multirow{5}{*}{3}                   & \textbf{55.1}    & 46.3   & 42.7   & 40.2   & 10.4 \\
    \cline{1-1}\cline{3-7}
    point transformer       &                                      & 57.2    & 51.2   & 49.4   & 48.3   & 11.7 \\
    \cline{1-1}\cline{3-7}
    no MFT                  &                                      & \multicolumn{4}{c|}{56.2}          & 7.5  \\
    \cline{1-1}\cline{3-7}
    MFT                     &                                      & 55.7    & \textbf{45.4}   & \textbf{40.9}   & \textbf{38.4}   & 10.1 \\
    \hline
    transformer & \multirow{5}{*}{5}                               & \textbf{54.9}    & 46.0  & 42.4    & 40.0   & 10.4 \\
    \cline{1-1}\cline{3-7}
    point transformer       &                                      & 57.0    & 50.9  & 49.2    & 48.2   & 11.7 \\
    \cline{1-1}\cline{3-7}
    no MFT              &                                      & \multicolumn{4}{c|}{56.0}          & 7.5  \\
    \cline{1-1}\cline{3-7}
    MFT                     &                                      & 55.4    & \textbf{45.1}  & \textbf{40.6}    & \textbf{38.2}   & 10.1 \\
    \hline
    transformer            & \multirow{5}{*}{7}                    & \textbf{54.7}    & 45.9  & 42.3    & 39.8   & 10.4 \\
    \cline{1-1}\cline{3-7}
    point transformer &                                            & 56.8    & 50.8  & 49.1    & 48.0   & 11.7 \\
    \cline{1-1}\cline{3-7}
    no MFT                     &                                       & \multicolumn{4}{c|}{55.7}          & 7.5  \\
    \cline{1-1}\cline{3-7}
    MFT            &                                       & 55.2    & \textbf{45.0}  & \textbf{40.5}    & \textbf{38.0}   & 10.1 \\
    \hline
  \end{tabular}
\end{table}

\begin{figure*}[t]
  \centering
  \includegraphics[width=1.0\textwidth]{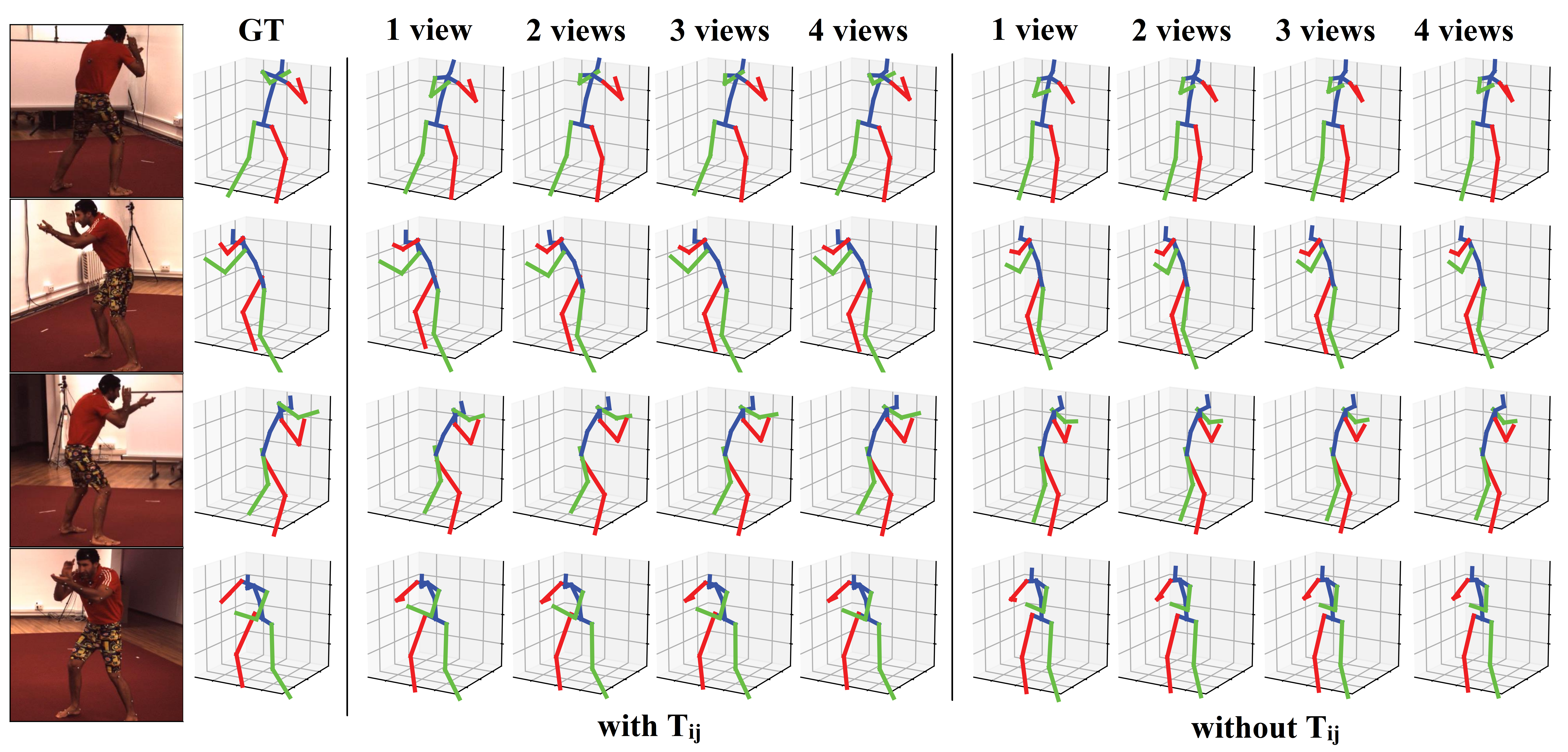}
  \caption{Demonstration of MFT with and without $T_{ij}$ on Human3.6M}\label{tran}
\end{figure*}

\begin{figure}[ht!]
  \centering
  \includegraphics[width=0.49\textwidth]{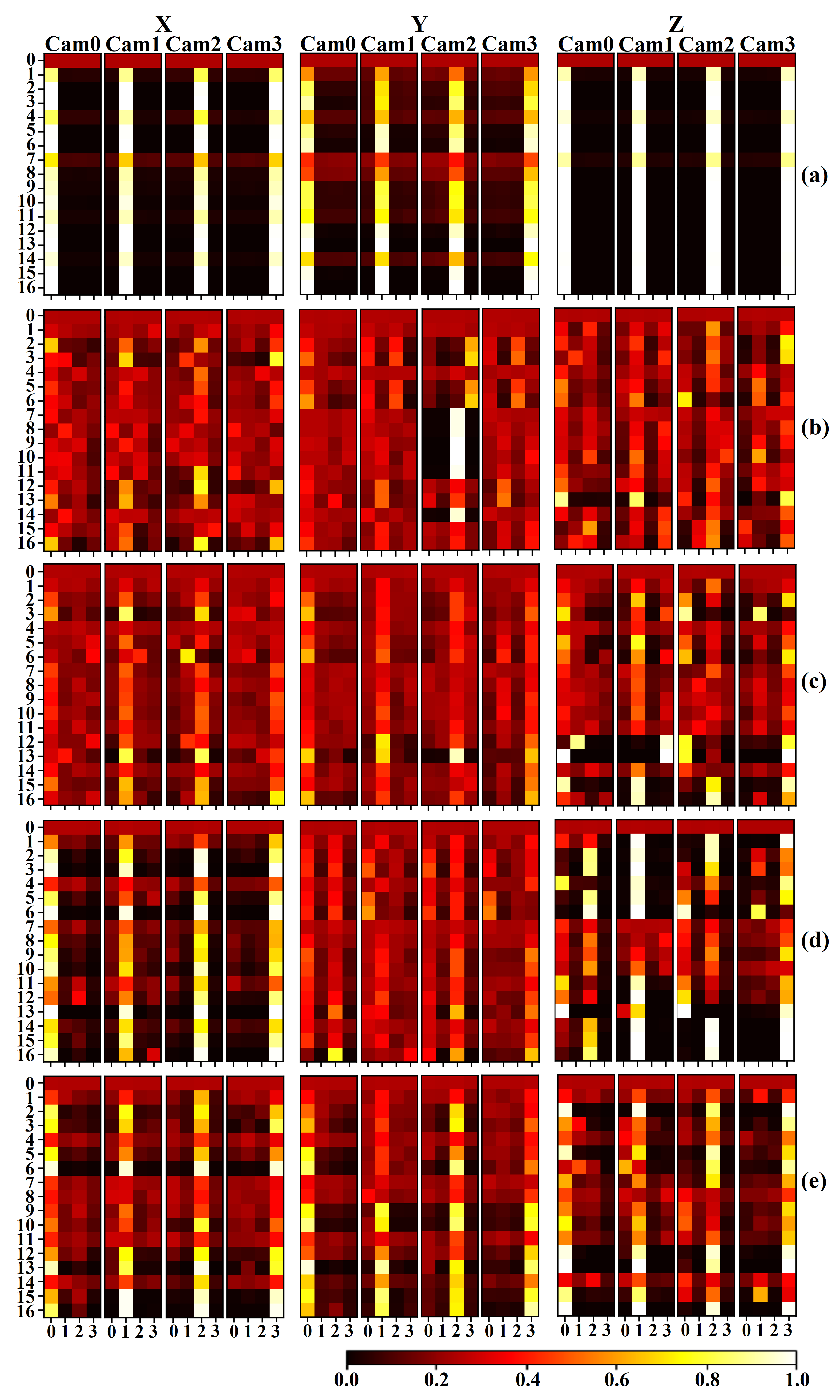}
  \caption{The contribution of the 2D pose (17 joint point) from 4 views to the 3D prediction (X, Y, Z coordinate) with different kind of Relative-Attention module. We measure the gradient of the predicted 3D coordinate to the features of each views and consider the maximum value as the contribution ratio. For better visualization, the values are normalized to the range of 0 to 1.
  (a) Transformer
  (b) Point Transformer trained on 4 views
  (c) MFT trained on 4 views
  (d) Point Transformer trained on 2 views
  (e) MFT trained on 2 views.}
  \label{vis_fusion}
\end{figure}

In the aspect of precision, we compare MFT with conventional transformer (removing absolution position encoding), point Transformer, and MFT without transform matrix $\mathbf{T}_{ij}$.
To adapt Point Transformer to our task, we replace the 3D coordinates of the point cloud with the flattened 2D pose from the 2D detector, resulting in more parameters to deal with relative position encoding.
Results in \tablename~\ref{relative_attention} demonstrate that MFT outperforms other methods in the vast majority of cases.
Notably, the performance of Point Transformer is only slightly inferior to MFT, reflecting the effectiveness of relative position encoding.
Besides, MFT and MFT w/o $\mathbf{T}_{ij}$ have little difference in results when only utilizing 1 view, but MFT achieves better performance when more viewpoints participate in multi-view fusing.
It means the transform matrix $\mathbf{T}_{ij}$ plays a vital role in multi-view feature fusing.
In addition, we display some results of MFT with and without $\mathbf{T}_{ij}$ in \figurename~\ref{tran}.
The MFT with $\mathbf{T}_{ij}$ can predict a more accurate 3D pose than the MFT with $\mathbf{T}_{ij}$, especially the position of hand and foot.
The reason is that $\mathbf{T}_{ij}$ can transform the feature from the source view to the target view. Then the transformed feature is fused with an element-wisely product.
Without $\mathbf{T}_{ij}$, the feature always lies in the source view, and the element-wise product is not effective enough for multi-view fusing.

In the aspect of generalization capability, we compare MFT with transformer and point Transformer.
We also report the performance of MTF-Transformer without MFT.
As MFT has the input and output of the same shape, removing MFT does not affect subsequent modules.
We train these models on two views (camera 0, 2) and test them in an increasing number of views from seen cameras (0, 2) to unseen cameras (1, 3).
As shown in \tablename~\ref{result_generalization}, the transformer achieves the best result when only 1 view is used.
MFT gets superior performance as the view number increases from 2 to 4.

\begin{table}[t]
  \centering
  \renewcommand{\arraystretch}{1.1}
  \caption{Results on Human3.6M with different setting of $D$. $T$ is the sequence length.}
    \begin{tabular}{c|c|c|c|c|c|c}
    \hline
    \multirow{2}{*}{D}& \multirow{2}{*}{$T$} & \multicolumn{4}{c|}{Number of views $N$} & \multirow{2}{*}{\begin{tabular}[c]{@{}c@{}}Parameter \\ (M)\end{tabular}}
    \\
    \cline{3-6}
                      &                              & 1     & 2    & 3    & 4     &  \\
    \hline
    1           & \multirow{6}{*}{1}                 & 51.8  & 37.0 & 31.7 & 29.6  & 10.074 \\
    \cline{1-1}\cline{3-7}
    2           &                                    & 51.3  & 35.9 & 30.6 & 28.4  & 10.076 \\
    \cline{1-1}\cline{3-7}
    3           &                                    & 51.2  & 35.9 & 30.7 & 28.5 & 10.079 \\
    \cline{1-1}\cline{3-7}
    4           &                                    & \textbf{50.7}  & \textbf{35.3} & \textbf{30.1} & \textbf{28.0}  & 10.083 \\
    \cline{1-1}\cline{3-7}
    5           &                                    & 50.9  & 35.8 & 30.6 & 28.4 & 10.088 \\
    \cline{1-1}\cline{3-7}
    6           &                                    & 51.3  & 35.5 & 30.3 & 28.1 & 10.095 \\
    \hline
    1           & \multirow{6}{*}{3}                 & 50.8  & 36.0 & 30.8 & 28.8 & 10.074 \\
    \cline{1-1}\cline{3-7}
     2           &                                   & 50.4  & 35.1 & 29.9 & 27.8 & 10.076 \\
    \cline{1-1}\cline{3-7}
    3           &                                    & 50.3  & 35.1 & 29.9 & 27.8 & 10.079 \\
    \cline{1-1}\cline{3-7}
    4           &                                    & \textbf{49.8}  & \textbf{34.5} & \textbf{29.4} & \textbf{27.3} & 10.083 \\
    \cline{1-1}\cline{3-7}
    5           &                                    & 50.1  & 35.0 & 29.9 & 27.8 & 10.088 \\
    \cline{1-1}\cline{3-7}
    6           &                                    & 50.4  & 34.7 & 29.6 & 27.5 & 10.095 \\
    \hline
    1           & \multirow{6}{*}{5}                 & 50.5  & 35.6 & 30.5 & 28.6 & 10.074 \\
    \cline{1-1}\cline{3-7}
    2           &                                    & 50    & 34.8 & 29.7 & 27.6 & 10.076 \\
    \cline{1-1}\cline{3-7}
    3           &                                    & 49.9  & 34.7 & 29.6 & 27.5 & 10.079 \\
    \cline{1-1}\cline{3-7}
    4           &                                    & \textbf{49.4}  & \textbf{34.2} & \textbf{29.2} & \textbf{27.1} & 10.083 \\
    \cline{1-1}\cline{3-7}
    5           &                                    & 49.7  & 34.7 & 29.7 & 27.6 & 10.088 \\
    \cline{1-1}\cline{3-7}
    6           &                                    & 50    & 34.4 & 29.4 & 27.3 & 10.095 \\
    \hline
    1           & \multirow{6}{*}{7}                 & 50.4  & 35.5 & 30.4 & 28.5 & 10.074 \\
    \cline{1-1}\cline{3-7}
    2           &                                    & 49.8  & 34.6 & 29.5 & 27.5 & 10.076 \\
    \cline{1-1}\cline{3-7}
    3           &                                    & 49.7  & 34.6 & 29.5 & 27.5 & 10.079 \\
    \cline{1-1}\cline{3-7}
    4           &                                    & \textbf{49.2}  & \textbf{34.1} & \textbf{29.1} & \textbf{27.1} & 10.083 \\
    \cline{1-1}\cline{3-7}
    5           &                                    & 49.5  & 34.6 & 29.6 & 27.5 & 10.088 \\
    \cline{1-1}\cline{3-7}
    6           &                                    & 49.8  & 34.2 & 29.3 & 27.2 & 10.095 \\
    \hline
    \end{tabular}%
  \label{dimention}%
\end{table}%
To further explain the utility of relative attention modules, we also display the contribution of the feature from each view to the final prediction in \figurename~\ref{vis_fusion}, inspired by the Grad-CAM\cite{selvaraju2017grad}.
There is a slight generalization gap when we train models on 4 views and test on the same number of views.
For the transformer, the feature from the target view makes almost the majority contribution to the final prediction.
Instead, MFT and point transformer get uniform contributions from all the views, indicating that relative position is essential to fuse multi-view information.
When we train MFT and point transformer on 2 views and test them on 4 views, there is a big generalization gap between the training and testing phase.
We can find that MFT fuses the feature from other views more effectively.
It is intuitive and verifies the effectiveness of the proposed Relative-Attention block.

In MFT, we divide the dimension $C$ of input $\boldsymbol{x}_i$ into $K$ groups.
To explore the effect of $K$ on the results, we train the model with different settings of $D$ because $C=D\times K$ and $D$ determines the shape of $\mathbf{T}_{ij}$ directly.
The results in \tablename~\ref{dimention} demonstrate MTF-Transformer achieves the best performance at the dimension of 4 with different sequence lengths.

\begin{table}[t]
  \centering
  \renewcommand{\arraystretch}{1.1}
  \caption{Results of different mask rate $M$ on Human3.6M. MTF-Transformer is trained on the training set with 5 views at different mask rate. We evaluate these models with different number of views as input. }
  \resizebox{0.48\textwidth}{!}{
  \setlength{\tabcolsep}{1.5mm}
  \begin{tabular}{c|c|cccccc}
    \hline
    \multicolumn{2}{c|}{Mask rate $M$}  & 0             & 0.2           & 0.4           & 0.6           & 0.8           & 1.0  \\
    \hline
    \multirow{4}{*}{\begin{tabular}[c]{@{}c@{}}Number of views \\ $N$\end{tabular}}
 &1                                  & 205.2         & 52.2          & 49.2          & 49.2  & \textbf{48.8}          & 49.4 \\
    \cline{2-8}
 &2                                  & 78.7         & 35.3          & \textbf{34.1} & 34.7          & 36.1          & 117.5 \\
    \cline{2-8}
 &3                                  & 44.5          & 29.3  & \textbf{29.1}           & 30.1          & 31.7          & 127.5 \\
    \cline{2-8}
 &4                                  & \textbf{25.6}  & 26.8          & 27.1          & 28.2          & 30.0          & 134.5 \\
    \hline
    \multicolumn{2}{c|}{Mean}           & 89.0         & 35.9          & \textbf{34.9} & 35.5          & 36.7          & 107.2 \\
    \hline
  \end{tabular}}
  \label{result_mask_rate}
\end{table}

\begin{figure}[t]
  \centering
  \includegraphics[width=0.48\textwidth]{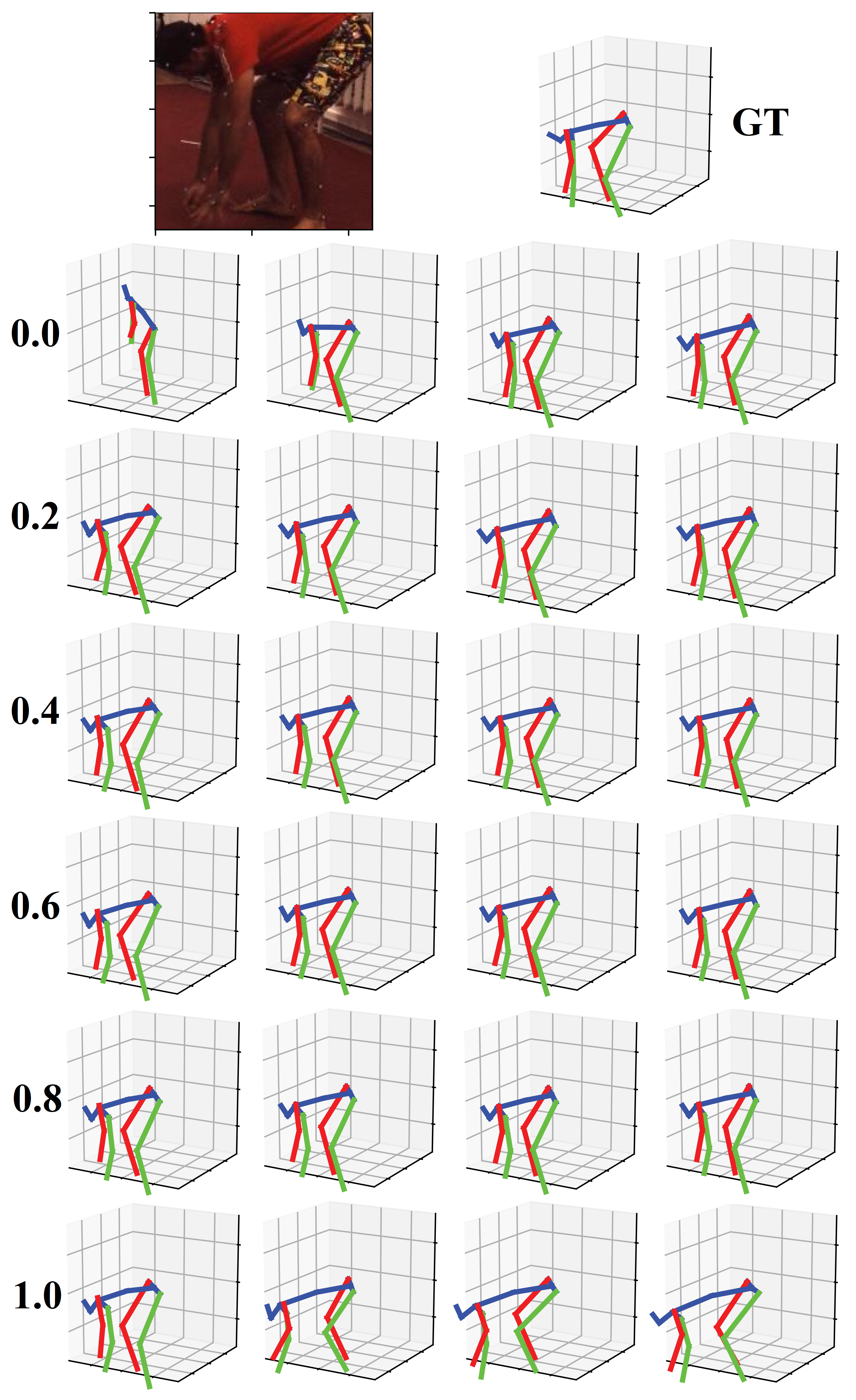}
  \caption{Some predictions under different mask rate}\label{vis_mask_rate}
\end{figure}

\subsubsection{Analysis on Random Block Mask}
Random Block Mask is designed to ensure the generalization capability of the MTF-Transformer.
To verify the effectiveness of the Random Block Mask, we train MTF-Transformer on Human3.6M training set with 5 views and set the mask rate $M$ at 0, 0.2, 0.4, 0.6, 0.8, 1.0, respectively.
With $M$ increasing, more features from different views are dropped in the training stage.
$M=0$ indicates that all the views participate in the feature fusing among all the views. Each view only fuses with itself when $M=1$.
In the testing stage, we test the MTF-Transformer counterparts with different mask rates via feeding testing samples with a different number of views (including 1, 2, 3, and 4 views).
The results are shown in \tablename~\ref{result_mask_rate}.
From the vertical comparison, at most mask rates, the performance of the MTF-Transformer gets better as the number of views increases, except for the mask rate of 1.
When the mask rate is set at 1, the MFT module fails to measure the relationship among the features since all the interconnections are masked.
It verifies that fusing multi-view features can improve the performance of 3D pose estimation.
From horizontal analysis, when the number of views is set at 4, MTF-Transformer achieves the best performance at the mask rate of 0. This number of views in the testing stage is close to that of the training stage (5 views).
As the number of views for testing decreases, the difference between training and testing is enlarged, and MTF-Transformer achieves the best performance at a higher mask rate.
It demonstrates that the Random Block Mask module is essential for scenes greatly different from the training stage.
The purpose of MFT-Transformer is to handle the input from an arbitrary number of views adaptively, so we evaluate the mean value of the MPJPE at different mask rates.
We find that the mask rate of 0.4 has the best result, and we will set the mask rate at 0.4 in all the experiments.
Some predict results are shown in \figurename~\ref{vis_mask_rate}.

\subsubsection{Analysis on Sequence Length}
\begin{table}[t]
  \centering
  \renewcommand{\arraystretch}{1.1}
  \caption{Results of different sequence length $T$ on Human3.6M.}
    \begin{tabular}{c|c|c|c|c}
    \hline
    \multirow{2}{*}{\begin{tabular}[c]{@{}c@{}}Sequence length \\ $T$\end{tabular}} & \multicolumn{4}{c}{Number of views $N$} \\
    \cline{2-5}
          & 1 view & 2 views& 3 views&4 views \\
    \hline
    1     & 51.62  & 35.13  & 29.74  & 27.46  \\
    3     & 50.70  & 34.28  & 28.99  & 26.77  \\
    5     & 50.28  & 33.91  & 28.69  & 26.51  \\
    7     & 50.01  & 33.71  & 28.55  & 26.41  \\
    9     & 49.82  & 33.57  & 28.44  & 26.32  \\
    11    & 49.66  & 33.47  & 28.38  & 26.28  \\
    13    & 49.53  & 33.42  & 28.35  & 26.27  \\
    15    & 49.45  & 33.39  & 28.35  & 26.27  \\
    17    & 49.39  & 33.37  & 28.34  & 26.28  \\
    19    & 49.33  & 33.34  & 28.33  & 26.27  \\
    21    & 49.27  & 33.32  & 28.32  & 26.26  \\
    23    & 49.23  & 33.29  & 28.30  & 26.24  \\
    25    & 49.21  & 33.27  & 28.28  & 26.22  \\
    27    & \textbf{49.19 } & \textbf{33.26 } & \textbf{28.27 } & \textbf{26.21 } \\
    \hline
    \end{tabular}%
  \label{result_tlen}%
\end{table}%
MTF-Transformer can adaptively handle videos with different sequence lengths.
We evaluate it via feeding videos with the length from 1 to 27. The results are shown in \tablename.~\ref{result_tlen}.
The performance of the MTF-Transformer increases as the sequence length increases, at any number of views as input.
It inflects that a more extended period of input benefits the pose estimation.
Interestingly, MTF-Transformer converges on certain precision as the sequence length number increases.
The more views are involved, MTF-Transformer converges on better precision and tends to saturate more quickly.
We utilize multi-view and temporal clues to estimate the pose of the middle frame under each viewpoint. Geometric and temporal information is complementary to each other.
Thus, when more multi-view clues are used, MTF-Transformer needs less temporal information to reconstruct the 3D pose.
Moreover, multi-view clues have some information that does not exist in temporal clues, so more viewpoints lead to better convergence results.

\subsubsection{Analysis on added synthesized views}
When training Human3.6M, we added a synthesized view to train MTF-Transformer as a data enhancement mechanism, following cheng et al.~\cite{cheng2019occlusion}.
To quantificat the effect of added views, we compare the results with a different number of added views in \tablename~\ref{addview}.
MTF-Transformer achieves the best performance when we add 1 synthesized view.
\begin{table}[t]
  \centering
  \renewcommand{\arraystretch}{1.1}
  \caption{Results on Human3.6M with different number of added view.}
  \renewcommand\arraystretch{1.0}
    \begin{tabular}{c|c|c|c|c|c|c}
    \hline
    \multirow{2}{*}{$T$} & \multirow{2}{*}{Added view} & \multicolumn{4}{c|}{The number of views $N$} &
    \multirow{2}{*}{\begin{tabular}[c]{@{}c@{}}Parameters\\(M)\end{tabular}} \\
    \cline{3-6}
                         &       & 1     & 2     & 3     & 4     &  \\
    \hline
    \multirow{3}{*}{1}   & 0 & 51.0  & 35.7 & 30.6 & 28.4 & \multirow{12}{*}{10.1} \\
    \cline{2-6}          & 1 & \textbf{50.7}  & \textbf{35.3} & \textbf{30.1} & \textbf{28.0} &  \\
    \cline{2-6}          & 2 & 51.4  & 35.8 & 30.4 & 28.1 &  \\
    \cline{1-6}    \multirow{3}{*}{3} & 0 & 50.2  & 35.0 & 30.0 & 27.9 &  \\
    \cline{2-6}          & 1 & \textbf{49.8}  & \textbf{34.5} & \textbf{29.4} & \textbf{27.3} &  \\
    \cline{2-6}          & 2 & 50.5  & 35.0 & 29.7 & 27.5 &  \\
    \cline{1-6}    \multirow{3}{*}{5} & 0 & 49.9  & 34.7 & 29.7 & 27.7 &  \\
    \cline{2-6}          & 1 & \textbf{49.4}  & \textbf{34.2} & \textbf{29.2} & \textbf{27.1} &  \\
    \cline{2-6}          & 2 & 50.1  & 34.8 & 29.5 & 27.2 &  \\
    \cline{1-6}    \multirow{3}{*}{7} & 0 & 49.7  & 34.6 & 29.6 & 27.6 &  \\
    \cline{2-6}          & 1 & \textbf{49.2}  & \textbf{34.1} & \textbf{29.1} & \textbf{27.1} &  \\
    \cline{2-6}          & 2 & 49.9  & 34.6 & 29.3 & \textbf{27.1 }&  \\
    \hline
    \end{tabular}%
  \label{addview}%
\end{table}%

\subsubsection{Analysis on computational complexity}
As shown in \tablename~\ref{computation}, we report the total number of parameters and estimated multiply-add operations (MACs) per frame (the 2D detector is not included).
For comparison, we also report parameters and MACs of Iskakov et al.~\cite{iskakov2019learnable}.
Similar to MTF-Transformer, Iskakov et al.~\cite{iskakov2019learnable} also infers the 3D pose via lifting multi-view 2D detections to 3D detections.
MTF-Transformer has a slightly less number of parameters and orders of magnitude less computational complexity.
The reason is that MTF-Transformer employs 1D convolution to manipulate the features instead of 3D convolution.
We also report the time consumption of MTF-Transformer in the training and testing phase in \tablename~\ref{time}.
MFT-Transformer is a magnitude faster than FLEX for inference time, tested in the same device.
Besides, increasing the number of views leads to a slight time consumption increment.
\begin{table}[t]
  \centering
  \renewcommand{\arraystretch}{1.1}
  \caption{Computational complexity of MTF-Transformer. We use THOP \protect\footnotemark to represent the number of parameters and MACs (multiply-add operations).
  $T$ is the length of sequence and $N$ is the number of views.}
    \begin{tabular}{c|c|c|c|c}
    \hline
    Method                           & $T$  & $N$ & Parameters(M)         & MACs(G) \\
    \hline
    Iskakov et al.~\cite{iskakov2019learnable}            & 1                  & 4                 & 11.9                  & 155     \\
    \hline
    FLEX~\cite{gordon2021flex}                   & 27                 & 4                 & 70.6                  & 4.27    \\
    \hline
    \multirow{8}{*}{MTF-Transformer} & \multirow{4}{*}{1} & 1                 & \multirow{4}{*}{10.1} & 0.01    \\
    \cline{3-3}\cline{5-5}
                                     &                    & 2                 &                       & 0.03    \\
    \cline{3-3}\cline{5-5}
                                     &                    & 3                 &                       & 0.05    \\
    \cline{3-3}\cline{5-5}
                                     &                    & 4                 &                       & 0.07    \\
    \cline{2-5}
                                     & \multirow{4}{*}{27}& 1                 & \multirow{4}{*}{10.1} & 0.27    \\
    \cline{3-3}\cline{5-5}
                                     &                    & 2                 &                       & 0.68    \\
    \cline{3-3}\cline{5-5}
                                     &                    & 3                 &                       & 1.23    \\
    \cline{3-3}\cline{5-5}
                                     &                    & 4                 &                       & 1.91    \\
    \hline
    \end{tabular}%
  \label{computation}%
\end{table}%

\begin{table}[t]
  \centering
  \renewcommand{\arraystretch}{1.1}
  \caption{The time consumption in training and testing phase.$T$ is the length of sequence and $N$ is the number of views.}\label{time}
  \begin{tabular}{c|c|c|c|c}
    \hline
    Methods & Device & $T$ & $N$ & time \\
    \hline
    \multicolumn{5}{c}{Training} \\
    \hline
    \hline
    \multirow{2}{*}{MTF-Transformer} & \multirow{2}{*}{2$\times$2080Ti} & 7  & - & 12h(60epochs) \\
    \cline{3-5}
                          &                                  & 27 & - & 34h(60epochs) \\
    \hline
    \hline
    \multicolumn{5}{c}{Testing} \\
    \hline
    FLEX~\cite{gordon2021flex}   & \multirow{9}{*}{1$\times$2080Ti} & 27 & 4 & 30.2ms \\
    \cline{1-1}\cline{3-5}
    \multirow{8}{*}{\begin{tabular}[c]{@{}c@{}}MTF-Transformer \end{tabular}}  &  & \multirow{4}{*}{1}  & 1 & 8.4ms \\
    \cline{4-5}
                                                                    &  &                     & 2 & 8.6ms \\
    \cline{4-5}
                                                                    &  &                     & 3 & 8.6ms \\
    \cline{4-5}
                                                                    &  &                     & 4 & 8.8ms \\
    \cline{3-5}
                                                                    &  & \multirow{4}{*}{27} & 1 & 9.2ms \\
    \cline{4-5}
                                                                    &  &                     & 2 & 9.3ms \\
    \cline{4-5}
                                                                    &  &                     & 3 & 9.5ms \\
    \cline{4-5}
                                                                    &  &                     & 4 & 9.9ms \\
    \hline
  \end{tabular}
\end{table}

\section{Conclusion}
We present a unified framework MTF-Transformer to fuse multi-view sequences in uncalibrated scenes with an arbitrary number of views.
MTF-Transformer can adaptively measure the relationship between each pair of views with a relative-attention mechanism, avoiding the dependency on camera calibration.
It is also computationally lightweight and can be directly applied to settings where the number of views and video frames varies.
Extensive experimental results demonstrate the effectiveness and robustness of the MTF-Transformer.

\section*{Acknowledgments}
This work was supported by the National Natural Science Foundation of China (61825601, U21B2044), and Science and Technology Program of Jiangsu Province (BK20192004B).

\footnotetext{\url{github.com/Lyken17/pytorch-OpCounter}}

\bibliographystyle{IEEEtran}
\bibliography{reference}

%

\begin{IEEEbiography}[{\includegraphics[width=1in,height=1.25in,clip,keepaspectratio]{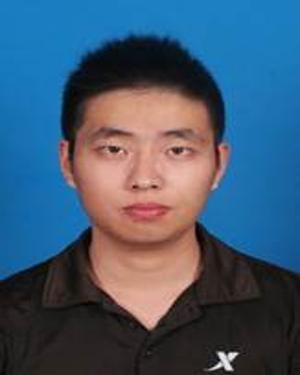}}]{Hui Shuai}
received the M.S degree from Nanjing University of Information Science and Technology (NUIST) in 2018.
He is currently working toward the PhD degree with NUIST.
His current research interests include object detection and 3D point cloud analysis.
\end{IEEEbiography}

\begin{IEEEbiography}[{\includegraphics[width=1in,height=1.25in,clip,keepaspectratio]{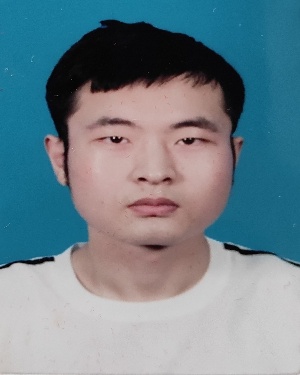}}]{Lele Wu}
received the bachelor's degree from the Nanjing University of Information Science and Technology (NUIST) in 2019, where he is currently pursuing the master's degree. His research interests include 3D human pose estimation and computer vision.
\end{IEEEbiography}

\begin{IEEEbiography}[{\includegraphics[width=1in,height=1.25in,clip,keepaspectratio]{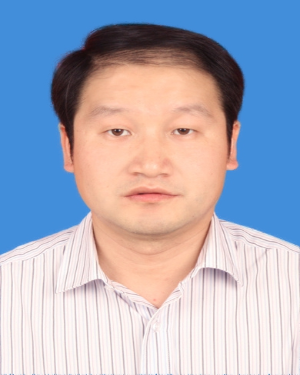}}]{Qingshan Liu (M’05–SM’07)}
received the M.S. degree from Southeast University, Nanjing, China, in 2000 and the Ph.D. degree from the Chinese Academy of Sciences, Beijing, China, in 2003.  He is currently a Professor in the School of Computer and Software, Nanjing University of Information Science and Technology, Nanjing. His research interests include pattern recognition, image and video analysis.
\end{IEEEbiography}




\end{document}